\def\ARXIVVERSION{1}
\documentclass{article}
\usepackage{iclr2026_conference,times}

\usepackage[hyphens]{url}
\usepackage{graphicx}
\usepackage{caption}
\usepackage{algorithm}
\usepackage{algorithmic}
\usepackage{amsmath}
\usepackage{amssymb}
\usepackage{bm}
\usepackage{booktabs}
\usepackage{colortbl}
\usepackage{pifont}
\usepackage{makecell}
\usepackage{multirow}
\usepackage{subcaption}
\usepackage{placeins}
\usepackage{wrapfig}
\definecolor{linkblue}{RGB}{120,170,255}

\newlength{\legacycolumnwidth}
\usepackage[
  colorlinks=true,
  linkcolor=black,
  citecolor=black,
  urlcolor=black,
  breaklinks=true
]{hyperref}

\ifdefined\ARXIVVERSION
  \iclrfinalcopy
\fi

\ifdefined\ARXIVVERSION
  \newcommand{\detailsref}[1]{Appendix~\ref{#1}}
\else
  \newcommand{\detailsref}[1]{the supplementary material}
\fi

\title{SCAIL-2: Unifying Controlled Character Animation with End-to-End In-Context Conditioning}
\author{
  Wenhao Yan\textsuperscript{1*}\quad
  Fengjia Guo\textsuperscript{1*}\quad
  Zhuoyi Yang\textsuperscript{1$\dagger$}\quad
  Jie Tang\textsuperscript{1$\ddagger$}\\
  \normalfont\textsuperscript{1}Tsinghua University\quad
  \textsuperscript{2}Zhipu AI
}

\begin{document}
\raggedbottom
\maketitle
\begingroup
\renewcommand{\thefootnote}{}
\footnotetext{\hspace*{-1.8em}\makebox[1.5em][l]{\textsuperscript{*}}Equal contribution.\\
\makebox[1.5em][l]{\textsuperscript{$\dagger$}}Tech lead.\\
\makebox[1.5em][l]{\textsuperscript{$\ddagger$}}Corresponding author.\\
Work done during internship at Zhipu AI.}
\addtocounter{footnote}{-1}
\endgroup

\begin{figure}[h]
  \centering
  \includegraphics[width=0.95\textwidth]{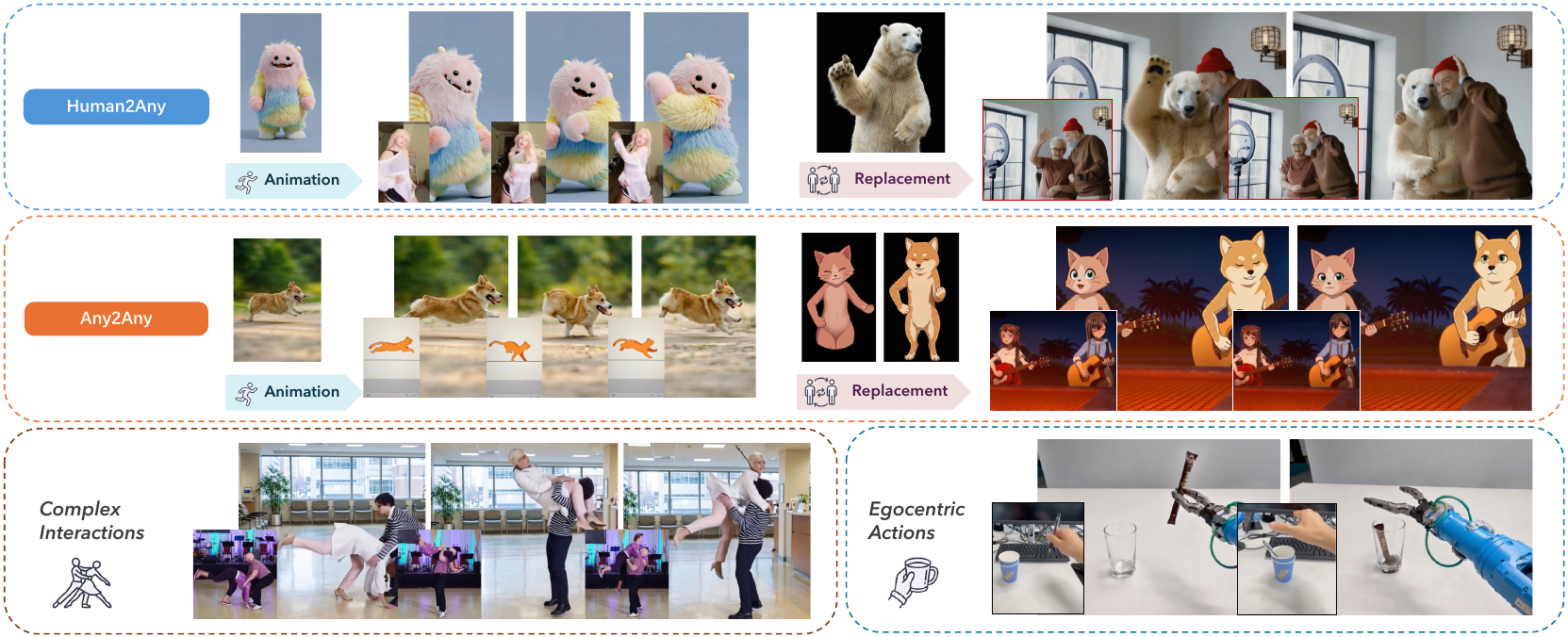}
  \caption{SCAIL-2 unifies various character animation tasks within an end-to-end paradigm.}
  \label{fig:teaser}
\end{figure}

\begin{abstract}
Controlled character animation aims to transfer motion from a driving sequence to a reference character. Prior works heavily rely on intermediate representations, such as pose skeletons for motion and masked backgrounds for environment, inevitably resulting in information loss. In this work, we present SCAIL-2, a framework that adopts an \textbf{end-to-end} driving paradigm by directly concatenating latent visual information to the model's input sequence. We enable end-to-end training through a data synthesis pipeline that produces MotionPair-60K, a curated dataset for several character animation subtasks. We unify the subtasks using decoupled conditions to accommodate different driving patterns, facilitated by In-Context Mask Conditioning and Mode-Specific RoPE, which provide soft guidance beyond textual instructions and visual information. To address synthetic discrepancy in detailed regions, we propose Bias-Aware DPO to construct preference items to mitigate the errors. Extensive experiments demonstrate that our method achieves state-of-the-art performance across various character animation tasks.
\ifdefined\ARXIVVERSION
Code, model weights, and a large subset of the dataset are available at
\href{https://teal024.github.io/SCAIL-2/}{\textcolor{linkblue}{\texttt{\textbf{https://teal024.github.io/SCAIL-2/}}}}.
\else
Code, model weights, and a large subset of the dataset will be made publicly available.
\fi

\end{abstract}

\section{Introduction}

\begin{wrapfigure}{r}{0.55\textwidth}
    \centering
    \includegraphics[width=0.98\linewidth]{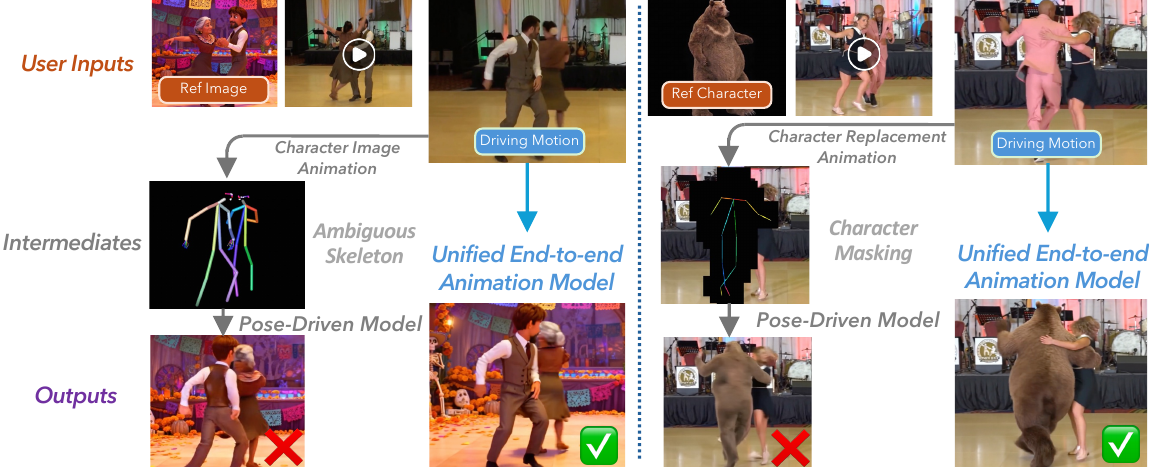}
    \caption{SCAIL-2 adopts an end-to-end driving paradigm, bypassing inaccurate poses and cropped backgrounds.}
    \label{fig:preteaser}
\end{wrapfigure}

Recent advances in video diffusion models (VDMs)~\citep{svd, cogvideox, wan} have substantially expanded the potential of controlled character animation~\citep{animateanyone, wananimate, scail} for film production and entertainment. Existing methods primarily rely on intermediate motion representations as conditions for VDMs to transfer motion. They typically use off-the-shelf pose estimators to draw skeleton maps~\citep{animateanyone, wananimate}, apply self-supervised bottlenecks~\citep{xunimotion, 3dmo}, or adopt test-time optimizations~\citep{followyourmotion, fastvmt} to obtain motion embeddings. Despite recent progress, skeleton maps remain inherently ambiguous in complex scenarios, as shown in Fig.~\ref{fig:preteaser}, while bottleneck encoders lack multi-character support and test-time methods struggle with detailed motions. Recent works~\citep{animatex, scail, onetoall} explore universal character animation for arbitrary characters, but still rely on human skeletal representations and struggle with out-of-distribution driving sources, such as using a cartoon cat to drive a dog in Fig.~\ref{fig:teaser}.

The reliance on such intermediates also introduces limitations in subtasks of character animation. Character replacement, typically defined as animation with environment affordance~\citep{animateanyone2}, is often framed as a pose-driven inpainting task using cropped background or objects as intermediates. The formulation is by design suboptimal, as changing the character may also affect the interacted objects and background. Furthermore, the character mask leaks the body shape and hinders cross-body-shape replacement. Another important subtask is multi-character animation, where pose-driven approaches suffer from misinterpretation of interaction when depth-ambiguous skeletons overlap. Existing methods for the task~\citep{dancetogether, hu2026multianimate} also apply masking to alleviate the issues, but do not fundamentally resolve the underlying ambiguity and further compromise body-shape adaptability.

We define \textbf{end-to-end character animation} as directly providing visual driving context rather than relying on information-losing intermediates. Training such a model, however, requires paired sequences where different characters perform the same motion, either within the same environment or across different environments. To address the scarcity of such data, we use pose-driven models during data construction to synthesize motion-aligned videos, employ an agentic editing loop for the generator to curate diverse high-quality animation pairs, and then reverse the resulting pairs for pose-free training.

Under this paradigm, we decouple the subtasks into end-to-end animation with task-specific conditions. To model their distinctions, we introduce \textbf{In-Context Mask Conditioning} and \textbf{Mode-Specific RoPE} as a unified interface under the reverse driving training paradigm. The in-context mask contains an Environment Switch that supports unification between vanilla animation and replacement, and further incorporates Binding Slots that describe motion–character binding to support single- and multi-character animation at the same time. This unification allows different subtasks to be composed within the same interface and share the optimization toward reasonable and coherent results. As a result, the model can address compositional tasks for which constructing dedicated data is difficult. A further challenge in end-to-end training is the bias of synthetic data, which we find most pronounced in detailed finger movements. We therefore design \textbf{Bias-Aware DPO}, a post-training scheme for better end-to-end motion capture in the hand regions.

Through end-to-end training, our model performs strongly across diverse scenarios, including complex motion, multi-character interactions, and non-human inputs for which pose estimators fail to capture detailed movements or fail entirely. The unified formulation further improves generalization in cross-identity motion transfer and environment integration. Our main contributions are summarized as follows:
\begin{enumerate}
    \item We propose an end-to-end conditioning paradigm to unify different tasks in character animation without lossy intermediate representations.
    \item We introduce a motion-pair synthesis pipeline to synthesize \textbf{MotionPair-60K}, a heterogeneous dataset to support end-to-end training.
    \item We apply a novel post-training mechanism, \textbf{Bias-Aware DPO}, to refine detailed end-to-end motion capture.
    \item We release \textbf{SCAIL-2}, an open-source pose-free animation model. Extensive experiments demonstrate that \textbf{SCAIL-2} achieves state-of-the-art performance across various animation tasks, unlocking emerging applications.
\end{enumerate}

\section{Related Works}
\noindent\textbf{Character Image Animation.}\enspace Character image animation refers to animating a character within its original background. Following Wan-Animate~\citep{wananimate}, we refer to this task as \textbf{\textit{Animation Mode}}. Existing pose-guided methods~\citep{animateanyone, champ, wananimate, everanimate} typically extract 2D skeletal motion sequences from the driving video and inject them into the video generation process. SCAIL~\citep{scail} introduces an identity-agnostic 3D skeleton representation rendered with different hues to enable multi-character animation, but the pose representation still discards visual information. The most relevant work on end-to-end animation is the closed-source DreamActor-M2~\citep{dreamactor-m2}, which aligns end-to-end driving capabilities with a pose-driven model. Our work instead trains directly on heterogeneous synthetic pairs to better utilize model priors and unify a wider range of subtasks, including complex interactions and character replacement.

\noindent\textbf{Character Replacement.}\enspace Character replacement animates a reference character within the environment of the driving video. Following Wan-Animate, we refer to this task as \textbf{\textit{Replacement Mode}}. Previous methods~\citep{animateanyone2, wananimate} formulate it as pose-driven animation with background inpainting. A recent advance, MoCha~\citep{mocha}, trains an end-to-end character replacement model using data rendered in Unreal Engine 5~\citep{ue5}. However, it struggles to generalize to characters with substantially different body shapes or to complex object interactions because of the limitations of renderer-generated training data. We address these limitations through end-to-end unification.

\section{Method}
\subsection{Preliminary}
\label{sec:preliminaries}
\noindent\textbf{General Task Formulation.}\enspace Given an input video $\boldsymbol{x}$, our flow-matching backbone~\citep{wan} first encodes it as $\boldsymbol{z}_0=\mathcal{E}(\boldsymbol{x})$ using a pretrained VAE encoder $\mathcal{E}$. We sample $\boldsymbol{z}_1\sim\mathcal{N}(0,\mathbf{I})$ and interpolate
\begin{equation}
\boldsymbol{z}_t=(1-t)\boldsymbol{z}_0+t\boldsymbol{z}_1,\quad
\boldsymbol{v}=\boldsymbol{z}_1-\boldsymbol{z}_0.
\end{equation}
A conditional velocity model $\boldsymbol{v}_\theta(\boldsymbol{z}_t,t,c)$ is trained with
\begin{equation}
\mathcal{L}_{\mathrm{FM}}=\mathbb{E}\!\left[
\|\boldsymbol{v}_\theta(\boldsymbol{z}_t,t,c)-\boldsymbol{v}\|_2^2\right],
\end{equation}
where $t\sim\mathcal{U}(0,1)$ and $c$ denotes auxiliary conditions.

For character animation, the condition $c$ comprises a text prompt $c_{\text{text}}$, a reference image $\boldsymbol{I}$ containing a character set $\mathcal{C}_{\boldsymbol{I}} = \{C_1, \ldots, C_N\}$ within an environment $E_{\boldsymbol{I}}$, and a motion signal derived from a driving video $\boldsymbol{y}$ containing characters $\mathcal{C}_{\boldsymbol{y}} = \{C_1^{\text{driv}}, \ldots, C_M^{\text{driv}}\}$ within environment $E_{\boldsymbol{y}}$.
Pose-driven methods first extract an explicit pose sequence $c_{\text{pose}} = \mathcal{P}(\boldsymbol{y})$ via an off-the-shelf estimator $\mathcal{P}$, then encode it into latent space as the motion condition.
In an end-to-end solution, the driving video is directly encoded, \textit{i.e.,} $\boldsymbol{z}_{\text{driv}} = \mathcal{E}(\boldsymbol{y})$, bypassing explicit pose estimation while still operating in the shared VAE latent space.

\noindent\textbf{Subtask Formulation.}\enspace We unify the subtasks of character image animation by a binding map $\pi: \mathcal{C}_{\boldsymbol{y}} \!\to\! \mathcal{C}_{\boldsymbol{I}}$ and an environment source $E \in \{E_{\boldsymbol{I}}, E_{\boldsymbol{y}}\}$.

\textit{Single Character Image Animation}: $E = E_{\boldsymbol{I}}$, $|\mathcal{C}_{\boldsymbol{y}}|=|\mathcal{C}_{\boldsymbol{I}}|=1$.

\textit{Multi Character Image Animation}: $E = E_{\boldsymbol{I}}$, $|\mathcal{C}_{\boldsymbol{y}}|=|\mathcal{C}_{\boldsymbol{I}}|>\!1$, where each driving character $C_i^{\text{driv}}$ transfers its motion to $\pi(C_i^{\text{driv}})$.

\textit{Character Replacement}: it shares the same binding formulation for both single- and multi-character scenarios, but takes $E = E_{\boldsymbol{y}}$.

We cast all subtasks as a single problem of reading different dimensions of information from the context and composing them into a plausible final result, and decompose the optimization into three objectives that the model should learn accordingly:

$\mathcal{O}_1$ \textit{Motion Binding} --- extract motions from the driving video while identifying their respective character origins, and route them solely to their bound targets $\pi(C_i^{\text{driv}})$;

$\mathcal{O}_2$ \textit{Environment Weaving} --- read the prescribed environment source $E$ from the context, and integrate the characters into the scene from either the reference ($E_{\boldsymbol{I}}$) or the driving video ($E_{\boldsymbol{y}}$) to generate a coherent composition;

$\mathcal{O}_3$ \textit{Universal Transfer} --- disentangle pose from identity to transfer motion from any driving character to any target in a physically plausible manner without identity leakage.

\setcounter{topnumber}{1}
\begin{table}[t]
  \centering
  \caption{Capability comparison with existing character animation methods. \textcolor{green!60!black}{\ding{51}}: Supported; \textcolor{orange!85!black}{\ensuremath{\bm{\triangle}}}: Performs poorly; \textcolor{red!75!black}{\ding{55}}: Unsupported.}
  \label{tab:capability-comparison}
  \small
  \setlength{\tabcolsep}{1pt}
  \renewcommand{\arraystretch}{1.12}
  \newcommand{\cmark}{\textcolor{green!60!black}{\ding{51}}}
  \newcommand{\xmark}{\textcolor{red!75!black}{\ding{55}}}
  \newcommand{\pmark}{\textcolor{orange!85!black}{\ensuremath{\bm{\triangle}}}}
  \begin{tabular}{@{}lcccc@{}}
    \toprule
    \multirow{2}{*}{\textbf{Method}}
      & \multicolumn{2}{c}{\textbf{Animation Mode}}
      & \multicolumn{2}{c}{\textbf{Replacement Mode}} \\
    \cmidrule(lr){2-3}\cmidrule(l){4-5}
      & \makecell{Multi-Character\\Interaction}
      & \makecell{Cross-Body-Shape}
      & \makecell{Simultaneous\\Multi-Character}
      & \makecell{Cross-Body-Shape} \\
    \midrule
    \makecell[l]{Wan-Animate~\citep{wananimate}}
      & \pmark & \pmark & \pmark & \pmark \\
    \makecell[l]{SteadyDancer~\citep{steadydancer}}
      & \pmark & \cmark & \xmark & \xmark \\
    \makecell[l]{Onetoall Animation\\\citep{onetoall}}
      & \pmark & \cmark & \xmark & \xmark \\
    \makecell[l]{SCAIL~\citep{scail}}
      & \cmark & \cmark & \xmark & \xmark \\
    \makecell[l]{MoCha~\citep{mocha}}
      & \xmark & \xmark & \xmark & \pmark \\
    \makecell[l]{DreamActor-M2~\citep{dreamactor-m2}}
      & \pmark & \cmark & \xmark & \xmark \\
    \makecell[l]{MultiAnimate~\citep{hu2026multianimate}}
      & \cmark & \pmark & \xmark & \xmark \\
    \makecell[l]{Kling 3.0 Motion Control\\\citep{klingmotion}}
      & \xmark & \cmark & \xmark & \xmark \\
    \rowcolor{black!7}
    \textbf{SCAIL-2 (Ours)}
      & \cmark & \cmark & \cmark & \cmark \\
    \bottomrule
  \end{tabular}
\end{table}
\FloatBarrier

\subsection{End-to-end Data Synthesis}
\begin{figure}[t]
  \centering
  \includegraphics[width=0.95\textwidth]{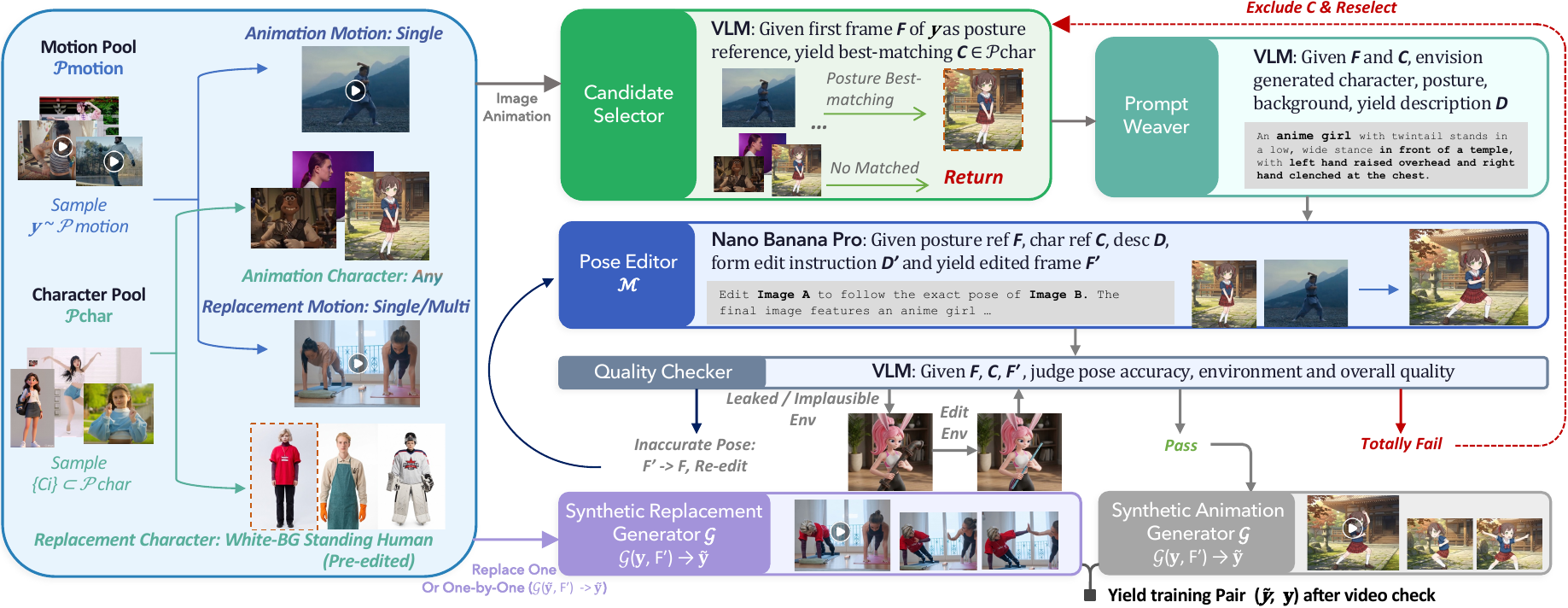}
  \caption{The overview of our synthetic pipeline for curating diverse high-quality cross-identity motion pairs.}
  \label{fig:pipeline}
\end{figure}

\noindent\textbf{Animation Synthetic Loop.}\enspace To achieve end-to-end character image animation, we need a synthetic engine to produce a synthetic video $\tilde{\boldsymbol{y}}$ from a ground-truth driving video $\boldsymbol{y}$ and a provided character reference image $\boldsymbol{I}$, through an animation generator $\mathcal{G}$:
\begin{equation}
\tilde{\boldsymbol{y}} = \mathcal{G}(\boldsymbol{y}, \boldsymbol{I}).
\end{equation}
Given a fixed $\mathcal{G}$ and a sampled driving sequence $\boldsymbol{y}$, our pipeline synthesizes optimal $\boldsymbol{I}$ for animation generators. To reduce synthetic cost, improve character diversity and reduce the generation of unreasonable data, we propose an agentic editing loop to generate plausible reference images directly from random human-centric datasets. The generation loop combines a Candidate Selector, a Prompt Weaver, a Quality Checker and a strong multi-reference image generation model $\mathcal{M}$~\citep{nanobanana}. Each iteration we directly sample one driving video and several character images. The Selector chooses the best character candidate, then provides $\mathcal{M}$ with the first frame of the video as the posture reference besides the character image. The Prompt Weaver is designed to pre-plan the desired character, background, and posture, bypassing context hallucination of $\mathcal{M}$'s innate planner and generator. We apply multiple turns of editing under the supervision of the Quality Checker to obtain better results. Additional editing is optionally applied to environment elements to prevent potential leakage and improve human-object-interaction (HOI) generalization.

For the choice of $\mathcal{G}$, we primarily use the pose-driven model SCAIL for cross-body-shape retargeting, supplemented by Wan-Animate for close-up shots. With the combination of our synthetic pipeline and the generator choice, we keep the discard rate of generated videos below 30\% when applying VLM~\citep{gemini} to check the synthetic data.

\newcommand{\evaluationtable}{%
\begin{table}[H]
  \centering
  \begin{minipage}[c]{0.48\textwidth}
  \centering
  \scriptsize
  \setlength{\tabcolsep}{2pt}
  \renewcommand{\arraystretch}{0.9}
  \begin{tabular}{@{}lccc@{}}
    \toprule
    \multirow{2}{*}{Method}
      & \multicolumn{3}{c}{X-Dance (Cross-Identity)} \\
    \cmidrule(l){2-4}
      & \makecell{Imaging\\Quality$\uparrow$}
      & \makecell{Temporal\\Consistency$\uparrow$}
      & \makecell{Appearance\\Consistency$\uparrow$} \\
    \midrule

    Wan-Animate
      & 3.80
      & 4.03
      & 4.23 \\

    Onetoall-Animation
      & 3.98
      & 3.99
      & 4.05 \\

    SteadyDancer
      & \underline{4.41}
      & 4.08
      & 4.17 \\

    SCAIL
      & 4.27
      & \textbf{4.21}
      & \underline{4.25} \\
    \midrule

    \textbf{SCAIL-2 (Ours)}
      & \textbf{4.43}
      & \underline{4.18}
      & \textbf{4.38} \\
    \bottomrule
  \end{tabular}
  \captionof{table}{Automatic metrics on X-Dance using the default driving paradigm for our method and baselines.}
  \label{tab:single-metrics}
  \label{tab:auto-eval}
  \end{minipage}\hfill
  \begin{minipage}[c]{0.48\textwidth}
  \centering
  \scriptsize
  \setlength{\tabcolsep}{1.5pt}
  \renewcommand{\arraystretch}{0.9}
  \begin{tabular}{@{}lcccc@{}}
    \toprule
    Method & SSIM$\uparrow$ & PSNR$\uparrow$ & LPIPS$\downarrow$ & FVD$\downarrow$ \\
    \midrule
    \textbf{SCAIL-2 (Ours)} & & & & \\
    \quad\textit{+ SAM3D-Body Mesh} & \textbf{0.6453} & \textbf{19.09} & 0.2231 & 287.11 \\
    \quad\textit{+ NLF Skeleton} & 0.6370 & 18.76 & 0.2285 & \textbf{282.85} \\
    SCAIL & & & & \\
    \quad\textit{+ SAM3D-Body Skeleton} & 0.6407 & 19.08 & \textbf{0.2212} & 309.63 \\
    \quad\textit{+ NLF Skeleton} & 0.6378 & 19.08 & \textbf{0.2212} & 312.79 \\
    Wan-Animate & 0.6340 & 18.62 & 0.2269 & 305.31 \\
    SteadyDancer & 0.6386 & 18.40 & 0.2311 & 332.20 \\
    Onetoall-Animation & 0.6138 & 17.25 & 0.2667 & 448.06 \\
    UniAnimate-DiT & 0.6367 & 18.52 & 0.2747 & 480.15 \\
    VACE & 0.5942 & 17.09 & 0.2883 & 387.52 \\
    HyperMotion & 0.6378 & 18.41 & 0.2339 & 411.30 \\
    \bottomrule
  \end{tabular}
  \captionof{table}{Single-character animation metrics on Studio-Bench's self-driven partition.}
  \label{tab:more-baselines}
  \end{minipage}
\end{table}
}

\noindent\textbf{Replacement Data.}\enspace We adopt a renderer-trained single-character replacement model~\citep{mocha} as the replacement generator and use only a small number of replacement pairs to supplement animation supervision: Multi-character animation pairs are hard to collect even with optimized models~\citep{scail, hu2026multianimate}, as the task complexity is significantly higher. We instead substitute multi-character animation with multi-character replacement, as replacement is more tractable and can be performed in a character-by-character manner. From the perspective of optimization objectives, the two subtasks mainly differ in $\mathcal{O}_2$ ($E_{\boldsymbol{I}}$ vs.\ $E_{\boldsymbol{y}}$) according to our formulation. Notably, the difference is already covered by single-character animation and replacement pairs, while the challenge of the multi-character setting, namely learning the binding $\pi$ for $\mathcal{O}_1$ and extracting and transferring motion under heavy inter-character occlusions for $\mathcal{O}_3$, is equally exercised by replacement.

\noindent\textbf{Reverse Driving.}\enspace The preceding synthesis pipeline constructs \textbf{MotionPair-60K}. These data are primarily used in a reverse manner: designated characters in a real video $\boldsymbol{y}$ are re-synthesized from $\boldsymbol{I}$ via pose transfer or one-by-one replacement, yielding a synthetic video $\tilde{\boldsymbol{y}}$. The synthetic $\tilde{\boldsymbol{y}}$ then serves as the driving input while the original real video $\boldsymbol{y}$ serves as the denoising target $\boldsymbol{x}$, alongside a reference frame $\boldsymbol{I}$ sampled from $\boldsymbol{y}$, to form a training triplet without introducing artifacts or renderer-bias by $\mathcal{G}$.

\subsection{Model Design}
\begin{figure}[t]
  \centering
  \includegraphics[width=0.95\textwidth]{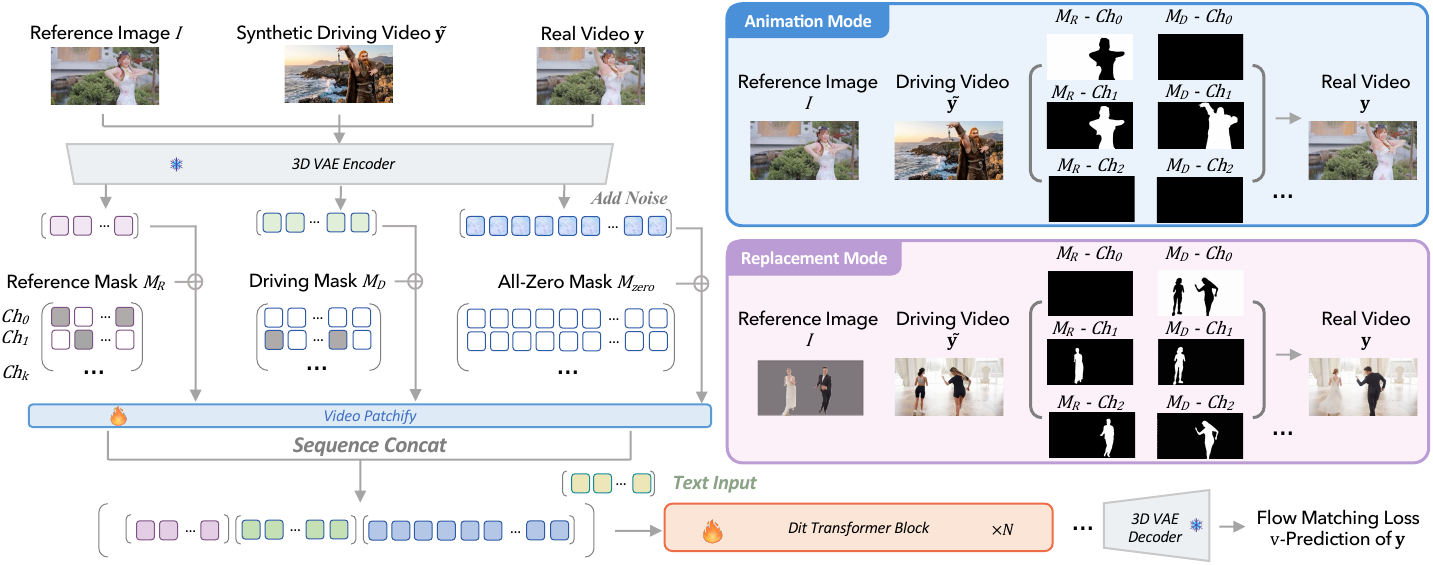}
  \caption{Overview of our model architecture and the context mask signal. $Ch_0$ means the mask channel for environment control, while $Ch_1$ to $Ch_K$ denote $K$ channels for character binding. The environment mask is the complement of the union of either the driving or the reference character masks. The model is adapted from the Wan2.1-14B-I2V backbone.}
  \label{fig:network}
\end{figure}

\noindent\textbf{Model Architecture.}\enspace Our model adopts the \textit{In-Context Driving} design from SCAIL, where the condition is concatenated to the denoised sequence rather than injected into the denoising embedding via channel concat~\citep{wananimate} or pose-guider~\citep{unianimate-dit}. Concretely, the I2V backbone~\citep{wan} receives the input of $[\boldsymbol{z}_{\text{ref}};\, \boldsymbol{z}_{\text{driv}};\, \boldsymbol{z}_{t}]$, the concatenation of the reference, driving video, and noisy video, where $\boldsymbol{z}_{\text{driv}}$ carries a spatial offset $\Delta_W$ along the $w$ axis so that it stays spatially detached from the video tokens.

\noindent\textbf{In-Context Mask Conditioning.}\enspace We propose In-Context Mask Conditioning to simultaneously model the difference between subtasks and enhance the original raw visual inputs for certain tasks, as shown in Fig.~\ref{fig:network}. To distinguish character image animation from replacement and optimize objective $\mathcal{O}_2$, we add $1$ additional channel as \textbf{Environment Switch}, indicating whether the environment should be derived from the reference image or the video. To optimize $\mathcal{O}_1$, we further introduce  $K$ channels as \textbf{Binding Slots}. The Binding Slots explicitly describe that the motion should flow exclusively within characters sharing a same channel. Single-character animation activates one random slot, while multi-character activates several. We set $K$ to 6 and randomly allow Binding Slots to be disabled under single-character settings.

In training, all valid mask signals are derived from reference images and driving sequences, without injecting signals from ground truth (the denoising latents keep an all-zero mask), which constitutes the fundamental distinction between our approach and prior works. The extraction is performed by a robust segment model SAM3~\citep{sam3} with its prompt-based matching. To align with the latent grid, the mask is downsampled spatially and stacked temporally along the channel dimension, producing $4(K{+}1)$ channels concatenated to the context. The introduction of such signals serves to provide enhanced guidance on top of the visual context to avoid confusion, rather than to alter it; the end-to-end nature is therefore preserved, as the model still observes the complete visual information.

\noindent\textbf{Mode-Specific Shifted RoPE.}\enspace To better model the differences between \textit{Animation Mode} and \textit{Replacement Mode}, we adopt Mode-Specific Shifted RoPE. We notice that \textit{Animation Mode} will regenerate a new starting frame from the visual elements in the reference, while \textit{Replacement Mode} needs an identical background from the first driving frame and only regenerates the character in the first frame. To model such difference, we design the denoising latent and reference in \textit{Animation Mode} with a temporal difference ($T=0$ and $T=1$), while \textit{Replacement Mode} uses a spatial difference, where an extra spatial RoPE shift $\Delta_H^{\text{ref}}$ is applied to $\boldsymbol{z}_{\text{ref}}$. For mode $m\in\{A,R\}$ and a video latent of shape $T_v\times H_v\times W_v$, the temporal, height, and width RoPE coordinate ranges are
\begin{equation}\label{eq:rope-coordinates}
\begin{aligned}
\boldsymbol{\rho}_{\rm ref}^{m}
  &=(0,\eta_m,[0,W_v)),\\
\boldsymbol{\rho}_{t}^{m}
  &=(\tau_m,[0,H_v),[0,W_v)),\\
\boldsymbol{\rho}_{\rm driv}^{m}
  &=(\tau_m,[0,H_v),[\Delta_W,\Delta_W{+}W_v)).
\end{aligned}
\end{equation}
For \textit{Animation Mode} ($m=A$), $\tau_A=[1,T_v]$ and $\eta_A=[0,H_v)$. For \textit{Replacement Mode} ($m=R$), $\tau_R=[0,T_v{-}1]$ and $\eta_R=[\Delta_H^{\rm ref},\Delta_H^{\rm ref}{+}H_v)$.

Both In-Context Mask Conditioning and RoPE differentiation are to prevent training conflict and allow the two tasks to share the optimization of universal target $\mathcal{O}_3$ and convey the trained universal capability to compositional tasks.

\subsection{Post Training}\label{sec:post-training}
\noindent\textbf{Bias-Aware DPO.}\enspace Although pose-free modeling enables the model to handle scenarios where pose extraction fails, the errors introduced by pose-driven synthetic data mean that the motion accuracy of end-to-end training data will be affected by pose estimation and animation generators. We notice that subtle movements in the hand region provide the most obvious evidence, where finger joints are often incorrectly articulated. To mitigate this bias, we propose Bias-Aware DPO. Specifically, we bootstrap synthetic data to directly simulate the errors introduced by pose estimators, and frame these errors as negative preferences for the model, thereby optimizing error correction during training.

\noindent\textbf{Preference Dataset Construction.}\enspace Our target is to construct positive--negative sample pairs that share the same reference identity and follow a consistent overall pose, but where the negative sample is always slightly less accurate than the positive one in fine-grained details. Given a motion video $y$, a pose estimator $P$, and an animation generator $\mathcal{G}$, we extract a pose $p = P(y)$ and synthesize two videos under different reference images $R$ and $S$:
\begin{equation}
    r = \mathcal{G}(p, R), \qquad s = \mathcal{G}(p, S).
\end{equation}
Sharing the same pose sequence, $r$ and $s$ form a basic pair: we take $s$ as the driving video and $r$ as the positive sample. The negative sample is then obtained by passing $r$ through one more round of error propagation along the same pipeline---re-extracting the pose from the synthesized video and regenerating under the same reference image:
\begin{equation}
    r^{-} = \mathcal{G}\big(P(r), R\big).
\end{equation}
where $r^{-}$ inherits one extra round of error and is therefore less accurate than $r$ in details. The gap can be further widened by performing the two extraction steps with $P'$ and $P''$, where we can select less accurate estimators as $P'$ or $P''$:
\begin{equation}\label{eq:negative-propagation}
    r^{-} = \mathcal{G}\big(P''(\mathcal{G}(P'(y), R)), R\big),
\end{equation}
To construct a preference data item, we randomly sample one frame from $r$ as the reference image $R_1$ and use $s$ as the driving video, forming a preference tuple:
\begin{equation}
    \left(s, R_1, r, r^{-}\right),
\end{equation}
where $(s, R_1)$ serves as the conditioning input, $r$ is the preferred sample, and $r^{-}$ the less preferred sample. With the preference tuple we can adopt DPO-based methods~\citep{Wallace_2024_diffusion_dpo} to optimize the preference. We provide further implementation details in \detailsref{app:post-training-details}.

\section{Experiments}
We full-finetune the backbone and perform DPO post-training after convergence. Detailed training recipes, evaluations, and ablations are supplemented in the Appendix.

\subsection{Evaluation Metrics}
We evaluate the performance of our methods using Studio-Bench~\citep{scail} and X-Dance~\citep{steadydancer}. To evaluate replacement performance, we select materials from Studio-Bench, supplement them with newly collected videos or images, and adapt them into cases that emphasize real-world cross-identity scenarios.
Cross-identity animation has no ground-truth target, so user studies provide the most reliable evaluation. Following Wan-Animate, we conduct blind Good/Same/Bad (GSB) user studies. We also employ Video-Bench~\citep{videobench} as the automatic evaluator for video quality following DreamActor-M2.

\subsection{Quantitative Evaluation}

\begin{figure}[H]
  \centering
  \includegraphics[width=\textwidth]{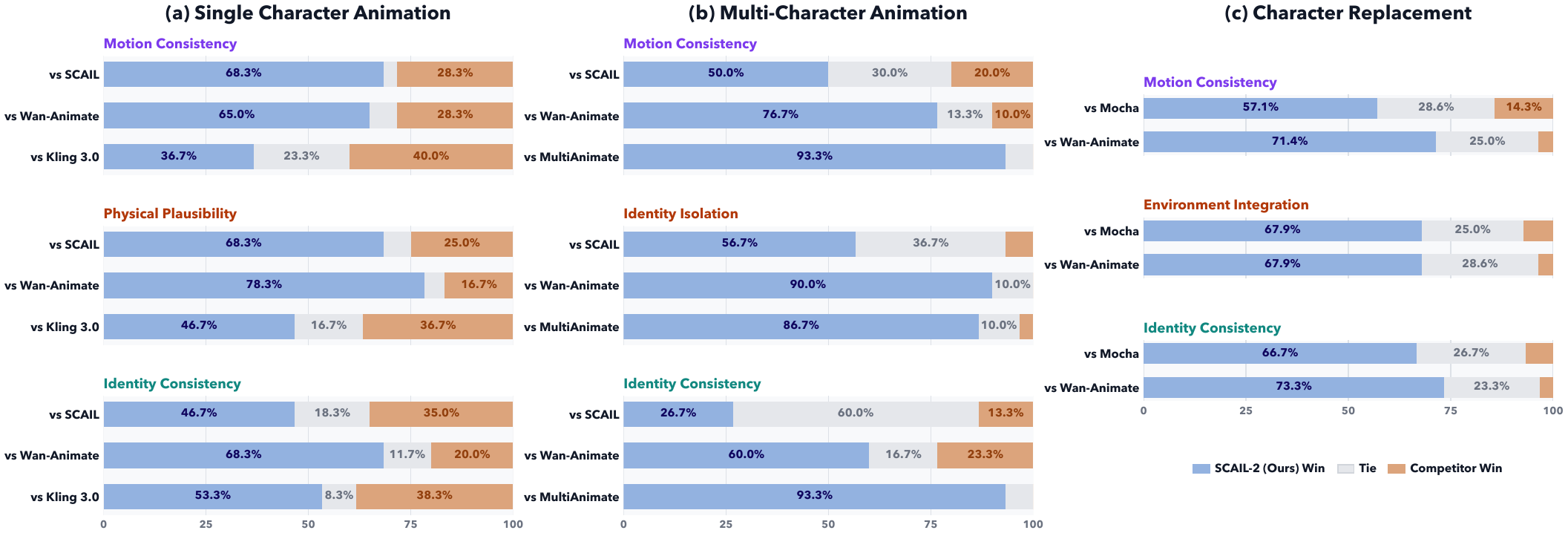}
  \caption{Human evaluation on Studio-Bench's cross-identity split: (a) single-character animation, (b) multi-character animation, and (c) character replacement. Kling 3.0 denotes Kling 3.0 Motion Control~\citep{klingmotion}.}
  \label{fig:human-eval}
\end{figure}

\evaluationtable

\noindent\textbf{Cross-Identity Results.}\enspace Results from Fig.~\ref{fig:human-eval}(a) show that for single-character animation our model wins over leading open-source works in human-evaluation metrics and remains close to proprietary service Kling 3.0. For multi-character animation our model demonstrates clear advantages even when the setting is zero-shot and the model learns from multi-replacement. Tab.~\ref{tab:single-metrics} also shows competitive video quality on X-Dance. In replacement mode (Fig.~\ref{fig:human-eval}(c)) our model is more preferred than inpainting-based Wan-Animate and our replacement data synthesizer MoCha, supporting the effectiveness of unification under our driving paradigm.

\noindent\textbf{Pose-driven Results.}\enspace As shown in Table~\ref{tab:more-baselines}, we compare against pose-driven methods Wan-Animate~\citep{wananimate}, SCAIL~\citep{scail}, UniAnimate-DiT~\citep{unianimate-dit}, VACE~\citep{vace}, and HyperMotion~\citep{hypermotion}. Though our model is not designed as a pose-driven generator and is trained with only a limited portion of pose-driven pairs, conditioning on human meshes from SAM3D-Body~\citep{sam3db} leads to competitive low-level metrics, even though the model has never seen this representation during training. We attribute the gain to the richer spatial information provided by precise meshes, which further demonstrates the ability of end-to-end animation to exploit information beyond sparse skeletal signals.

\FloatBarrier

\subsection{Qualitative Evaluation}

\begin{figure}[H]
  \centering
  \includegraphics[width=\textwidth]{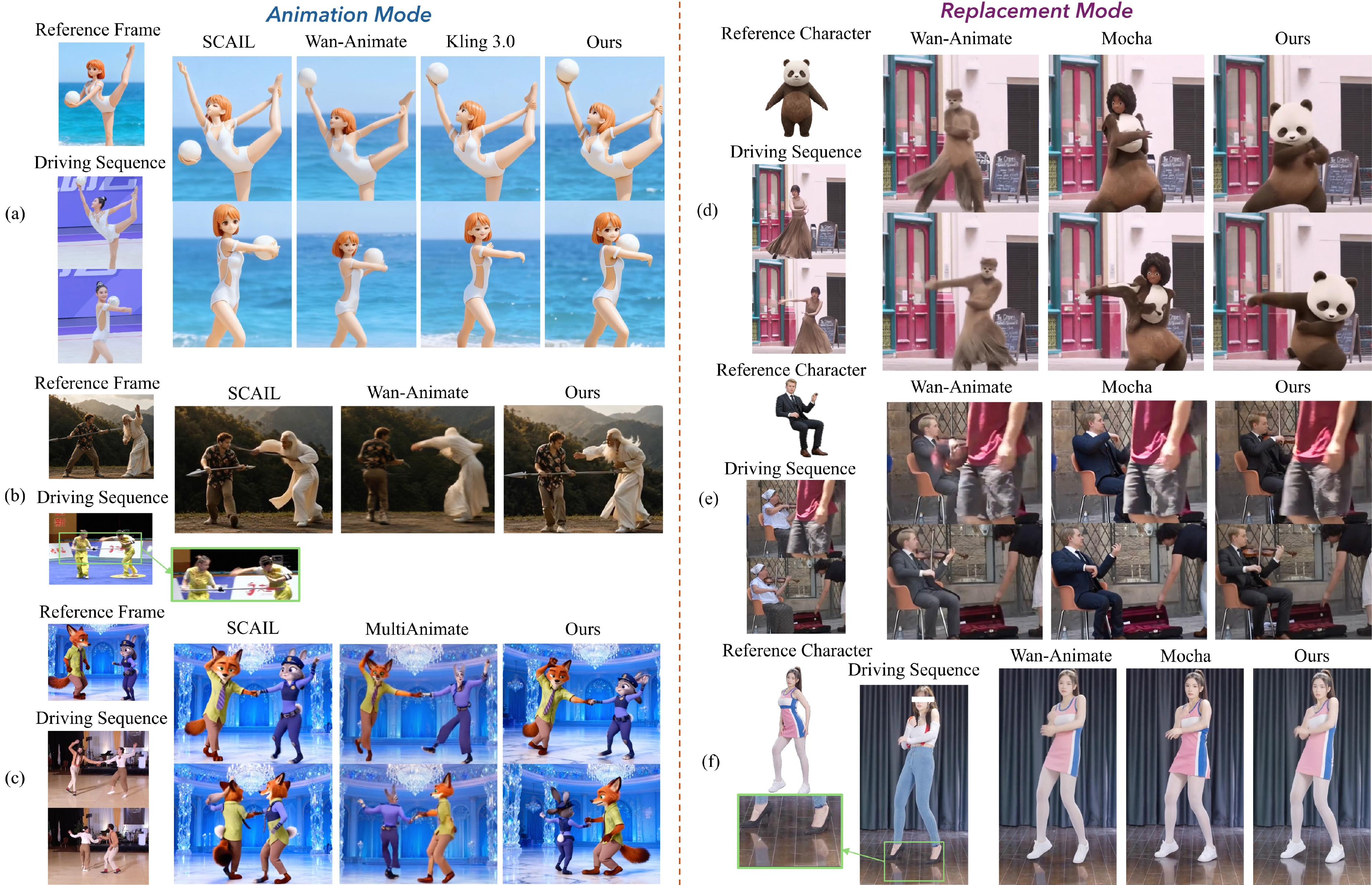}
  \caption{Qualitative comparison against baselines under cross-identity inputs.}
  \label{fig:compare}
\end{figure}

Fig.~\ref{fig:compare} presents qualitative comparisons with SoTA baselines under cross-identity inputs. Our model produces accurate motion while preserving identity and fine-grained human--object interactions. Figures~\ref{fig:compare}(a) and~\ref{fig:compare}(b) show challenging cases in which the baselines fail to reproduce intricate movements involving object interactions. Figure~\ref{fig:compare}(c) demonstrates identity isolation and body-shape consistency in a multi-character setting.

For \textit{replacement mode}, Fig.~\ref{fig:compare}(d) shows that our model preserves both motion accuracy and character generalization. Figure~\ref{fig:compare}(e) requires maintaining the character's identity and hand--instrument interaction amid a crossing crowd; MoCha loses the instrument, while Wan-Animate produces dark artifacts around the person because of its inpainting mechanism. In Fig.~\ref{fig:compare}(f), the shoe reflections further show that our model integrates the character most naturally into the environment.

\noindent
\begin{minipage}[t]{0.52\textwidth}
\vspace{0pt}
\textbf{Zero-Shot Abilities.}\enspace The teaser in Fig.~\ref{fig:teaser} demonstrates that our method can generalize to cartoon animal inputs and egocentric inputs in a completely zero-shot manner. By learning visual motion transfer directly within the DiT, our formulation can leverage its broad pretrained priors instead of restricting motion to human-centric intermediate representations. The results also suggest that the learned capability captures a form of general motion transfer beyond conventional third-person character animation. We further validate this observation on object motion, which differs substantially from the human sports and dance patterns that dominate the training data.
\end{minipage}\hfill
\begin{minipage}[t]{0.44\textwidth}
\vspace{0pt}
    \centering
    \includegraphics[width=0.90\linewidth]{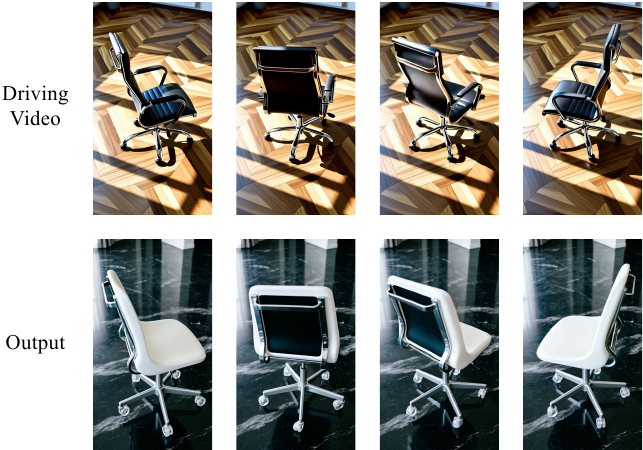}
    \addtocounter{figure}{1}
    \captionof{figure}{SCAIL-2 exhibits an emergent zero-shot ability to transfer object motion.}
    \label{fig:object}
\end{minipage}

\noindent Despite being trained exclusively on human motion, our method exhibits an emergent ability to follow object motion, as shown in Fig.~\ref{fig:object}: in the driving video, the chair first rotates counterclockwise, then abruptly reverses to clockwise rotation, before continuing to rotate counterclockwise, and the model's output faithfully follows this motion.

\FloatBarrier
\subsection{Ablation Studies}

\noindent\textbf{Ablation on Driving Modes.}\enspace Figure~\ref{fig:ablation}(a) compares different driving modes of our model. Replacing the original visual driving signal with skeletons degrades motion fidelity. Together with the comparisons against other pose-driven methods, this result demonstrates the benefit of the end-to-end visual driving paradigm.

\noindent\textbf{Ablation on Network Modules.}\enspace Fig.~\ref{fig:ablation}(b) demonstrates the effectiveness of Binding Slots mask. Inference without the character mask fails to maintain identity when pedestrians pass through; training without Binding Slots forces the model into innate tracking but affects the pedestrian's outfit. The slots also help steady identity assignment in rotating scenarios, indicating an additional mask signal remains important even atop an end-to-end formulation.
Fig.~\ref{fig:ablation}(c) shows that both the Environment Switch and Mode-Specific RoPE are necessary for unifying animation and replacement: removing the former leads to abnormal backgrounds, while removing the latter makes generation susceptible to certain patterns (like dark area) in the reference image.

\label{app:quan-abl}
\noindent
\begin{minipage}[t]{0.43\textwidth}
\vspace{0pt}
We further provide quantitative ablations under the cross-identity multi-character animation part of Studio-Bench. 
As shown in Table~\ref{tab:ablation-metrics}, removing Binding Slots causes a clear drop in Appearance Consistency, demonstrating that the slots are key to keeping each character's identity intact.
\end{minipage}\hfill
\begin{minipage}[t]{0.53\textwidth}
  \vspace{0pt}
  \centering
  \scriptsize
  \setlength{\tabcolsep}{2.5pt}
  \begin{tabular}{@{}l|ccc@{}}
    \toprule
    \multirow{2}{*}{\textbf{Method}} & \multicolumn{3}{c}{Studio-Bench's multi-character split} \\
    \cmidrule(l){2-4}
      & \makecell{Imaging\\Quality $\uparrow$}
      & \makecell{Temporal\\Consistency $\uparrow$}
      & \makecell{Appearance\\Consistency $\uparrow$} \\
    \midrule
    \textit{w/o} Binding Slots & 4.47 & 4.17 & 3.90 \\
    \textit{w/o} Replacement & 3.90 & 4.13 & 4.10 \\
    \midrule
    \textbf{Ours (Full)} & \textbf{4.63} & \textbf{4.23} & \textbf{4.13} \\
    \bottomrule
  \end{tabular}
  \captionof{table}{Quantitative ablations of Binding Slots and Replacement Data under multi-character animation.}
  \label{tab:ablation-metrics}
\end{minipage}

\noindent\textbf{Ablation on Data Composition.}\enspace Fig.~\ref{fig:ablation}(d) and (e) demonstrate the synergistic effect of unification. Without replacement data, the model fails to extract correct motion when characters overlap; without animation data, the model struggles to maintain motion consistency when body shape changes drastically. As shown in Table~\ref{tab:ablation-metrics}, replacement data contributes positively to multi-character animation.

\begin{figure}[!t]
  \centering
  \includegraphics[width=\textwidth]{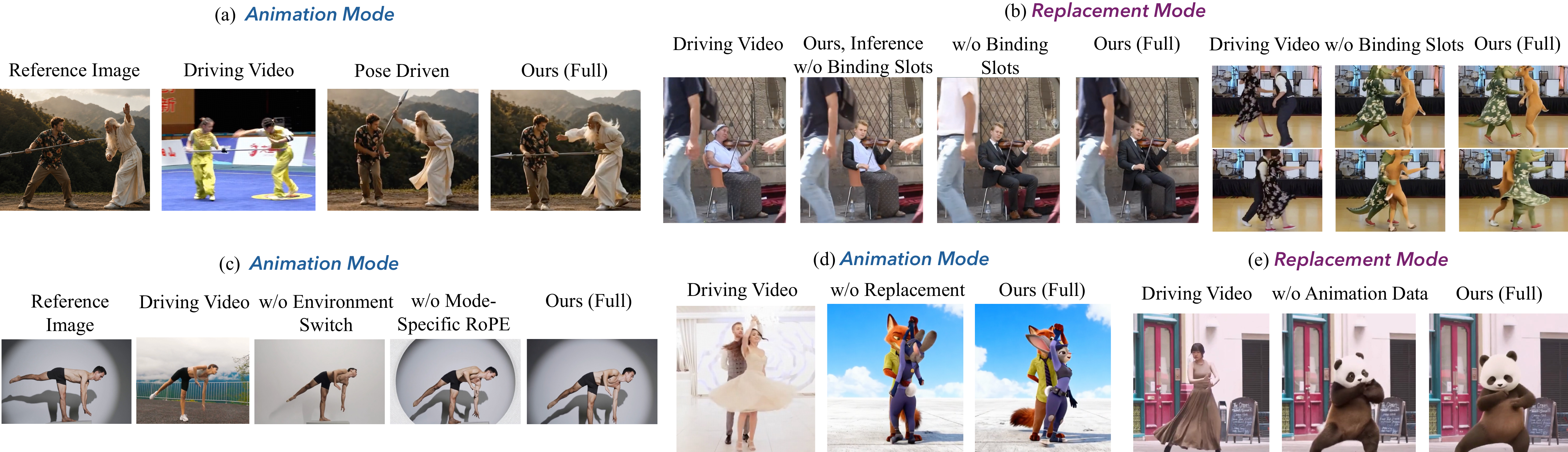}
  \setcounter{figure}{6}
  \caption{Ablation studies on modules and data. Zoom in for better details.}
  \label{fig:ablation}
\end{figure}
\setcounter{figure}{8}

\begin{figure}[H]
  \centering
  \begin{minipage}[c]{0.48\textwidth}
    \centering
    \includegraphics[width=0.70\linewidth]{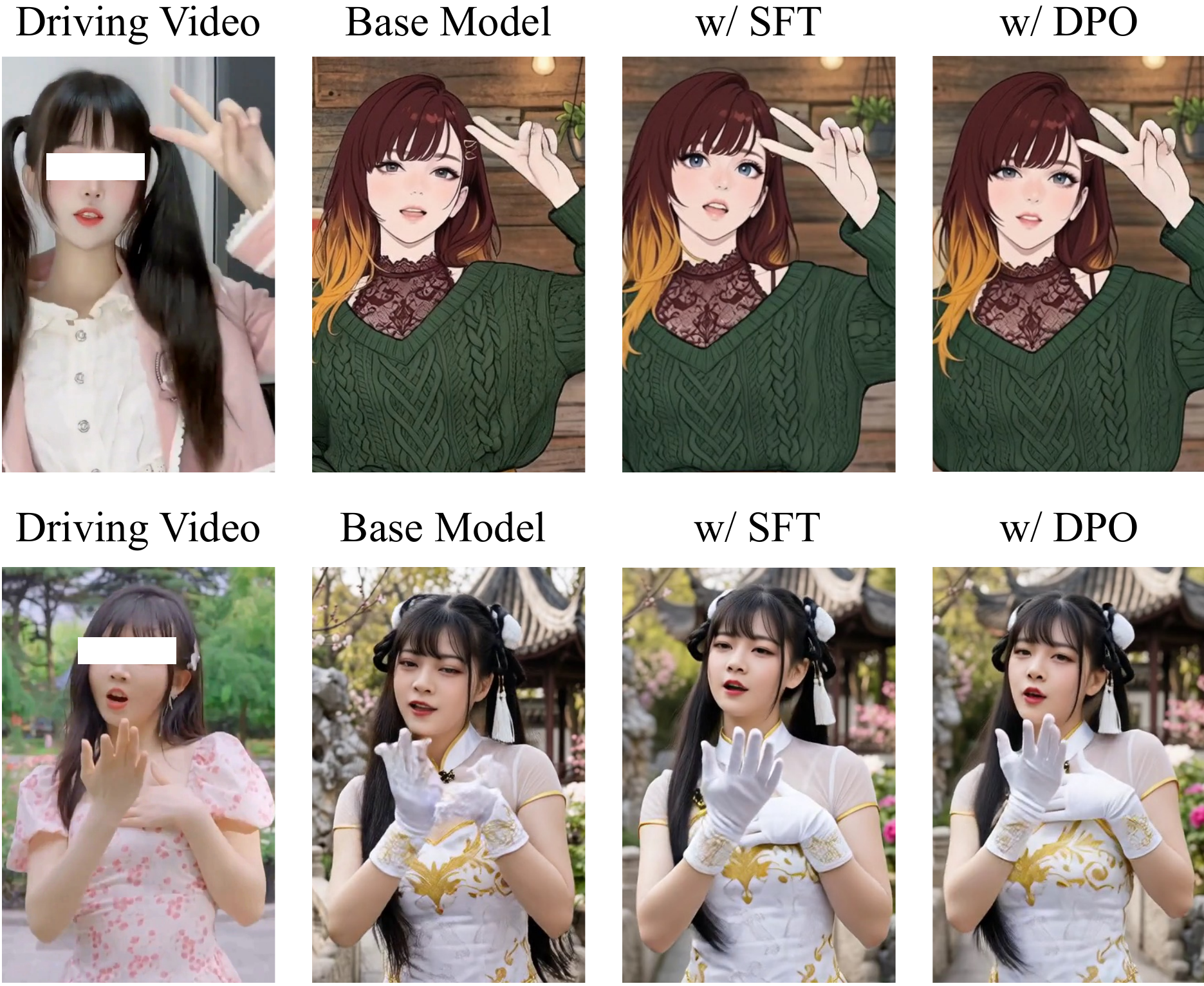}
    \captionof{figure}{Qualitative ablation on our Bias-Aware DPO.}
    \label{fig:dpo}
  \end{minipage}\hfill
  \begin{minipage}[c]{0.48\textwidth}
    \centering
    \includegraphics[width=0.8\linewidth]{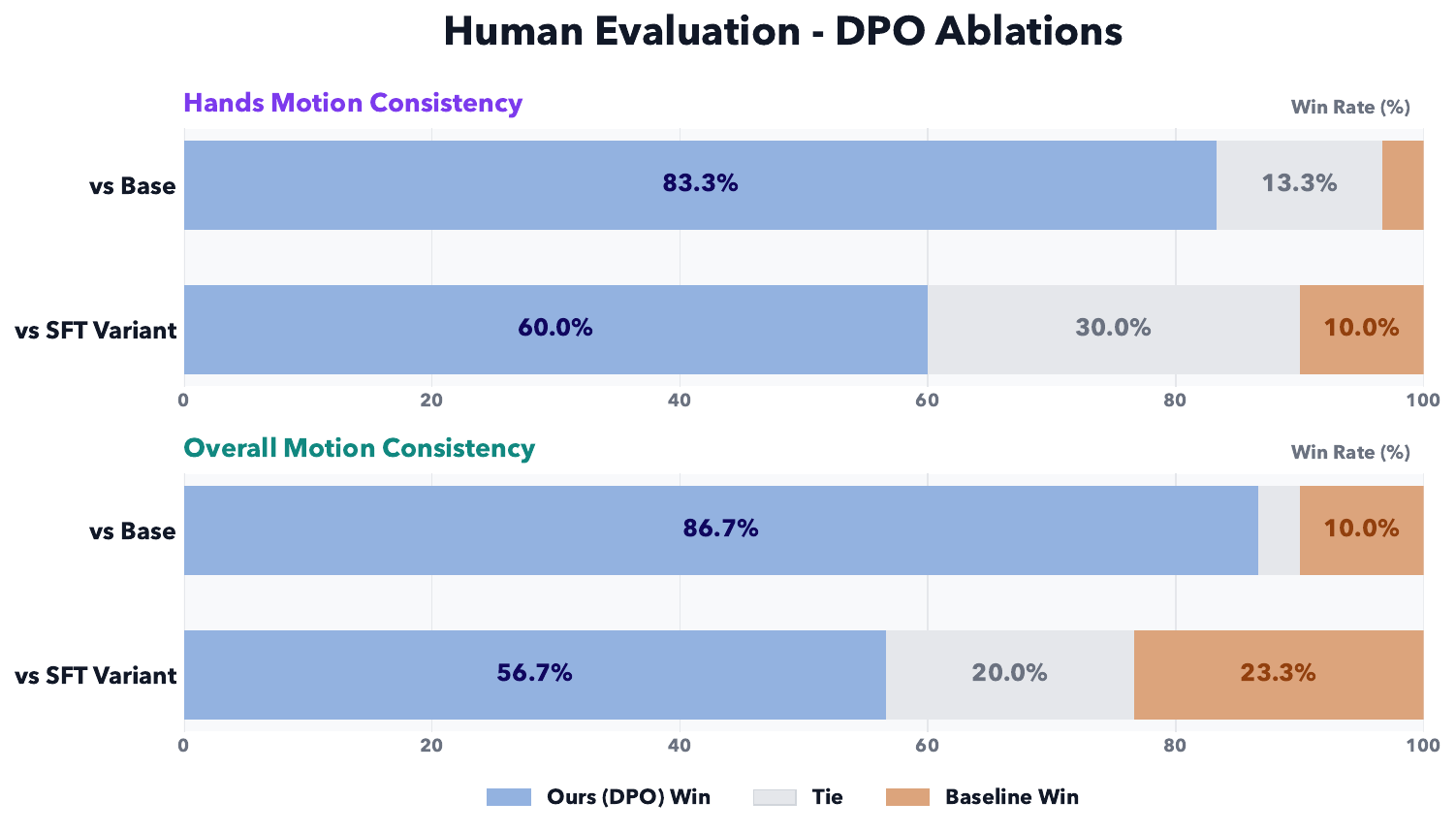}
    \captionof{figure}{Quantitative ablation on our Bias-Aware DPO.}
    \label{fig:dpo_ablation}
  \end{minipage}
\end{figure}

\noindent\textbf{Ablations on Bias-Aware DPO.}\enspace Figure~\ref{fig:dpo} compares Bias-Aware DPO, weighted SFT, and the base model. Weighted SFT emphasizes hand regions through an explicit regional loss, but its optimization remains limited without negative samples. In contrast, Bias-Aware DPO explicitly models synthetic errors and captures finer hand details. We further conduct quantitative ablations under our DPO training on the single-character cross-identity split of Studio-Bench, as shown in Fig.~\ref{fig:dpo_ablation}. Enabling the DPO LoRA improves hand motion consistency and is consistently preferred over the SFT variant.

Although the DPO loss is computed only within hand regions, it updates the policy globally rather than locally: the hand mask merely up-weights the most salient errors instead of confining optimization to the masked region. As shown in Fig.~\ref{fig:dpo}, this leads to visible improvements on other fine details such as the mouth. The subjective comparison in Fig.~\ref{fig:dpo_ablation} shows that Bias-Aware DPO is preferred both in hands and overall motion consistency, compared with weighted SFT which similarly emphasizes the hands but only fits positive samples.

\FloatBarrier
\section{Limitations}
While end-to-end designs feed the model complete visual information that is naturally richer, the fundamental limitation lies in a strict dependence on large-scale, high-quality paired training data. While our synthetic pipeline largely resolves the data-scarcity problem, the fidelity of the constructed data still hinges on the capability of these generators. We use Bias-Aware DPO to model the preference against bias, but reliable positive samples for fine-grained regions remain hard to obtain. Future work could adopt more advanced models to synthesize higher-quality data and extend the framework to more tasks, such as accurate lip-syncing for talking scenes.

\section{Conclusions}
In this work, we present \textbf{SCAIL-2}, an end-to-end framework for character animation. We curate a pipeline that synthesizes an end-to-end dataset spanning diverse animation tasks, making end-to-end animation feasible at scale. By training a DiT with strong priors to directly extract and convert information from visual contexts, the model generalizes to a broader range of zero-shot inputs. Building on this end-to-end paradigm, we unify several subtasks and observe a clear synergistic effect. A novel Bias-Aware DPO post-training scheme is further introduced to effectively mitigate regional synthetic bias. Extensive experiments demonstrate that SCAIL-2 achieves state-of-the-art performance across input conditions. We believe SCAIL-2 offers a practical and extensible paradigm towards production-ready character animation.

\clearpage
\ifdefined\PROJECTROOTBUILD
  \bibliography{AnonymousSubmission/LaTeX/references}
  \bibliographystyle{AnonymousSubmission/LaTeX/iclr2026_conference}
\else
  \bibliography{references}

\begin{thebibliography}{42}
\providecommand{\natexlab}[1]{#1}
\providecommand{\url}[1]{\texttt{#1}}
\expandafter\ifx\csname urlstyle\endcsname\relax
  \providecommand{\doi}[1]{doi: #1}\else
  \providecommand{\doi}{doi: \begingroup \urlstyle{rm}\Url}\fi

\bibitem[Blattmann et~al.(2023)Blattmann, Dockhorn, Kulal, Mendelevitch,
  Kilian, Lorenz, Levi, English, Voleti, Letts, et~al.]{svd}
Andreas Blattmann, Tim Dockhorn, Sumith Kulal, Daniel Mendelevitch, Maciej
  Kilian, Dominik Lorenz, Yam Levi, Zion English, Vikram Voleti, Adam Letts,
  et~al.
\newblock Stable video diffusion: Scaling latent video diffusion models to
  large datasets.
\newblock \emph{arXiv preprint arXiv:2311.15127}, 2023.

\bibitem[Carion et~al.(2025)Carion, Gustafson, Hu, Debnath, Hu, Suris, Ryali,
  Alwala, Khedr, Huang, et~al.]{sam3}
Nicolas Carion, Laura Gustafson, Yuan-Ting Hu, Shoubhik Debnath, Ronghang Hu,
  Didac Suris, Chaitanya Ryali, Kalyan~Vasudev Alwala, Haitham Khedr, Andrew
  Huang, et~al.
\newblock Sam 3: Segment anything with concepts.
\newblock \emph{arXiv preprint arXiv:2511.16719}, 2025.

\bibitem[Chen et~al.(2025)Chen, Chen, Xu, Li, Dong, Sun, Jiang, Li, Yang, Zhao,
  et~al.]{dancetogether}
Junhao Chen, Mingjin Chen, Jianjin Xu, Xiang Li, Junting Dong, Mingze Sun,
  Puhua Jiang, Hongxiang Li, Yuhang Yang, Hao Zhao, et~al.
\newblock Dancetogether! identity-preserving multi-person interactive video
  generation.
\newblock \emph{arXiv preprint arXiv:2505.18078}, 2025.

\bibitem[Cheng et~al.(2025)Cheng, Gao, Hu, Hu, Huang, Ji, Li, Meng, Qi, Qiao,
  et~al.]{wananimate}
Gang Cheng, Xin Gao, Li~Hu, Siqi Hu, Mingyang Huang, Chaonan Ji, Ju~Li, Dechao
  Meng, Jinwei Qi, Penchong Qiao, et~al.
\newblock Wan-animate: Unified character animation and replacement with
  holistic replication.
\newblock \emph{arXiv preprint arXiv:2509.14055}, 2025.

\bibitem[Contributors(2025)]{lightx2v}
LightX2V Contributors.
\newblock Lightx2v: Light video generation inference framework.
\newblock \url{https://github.com/ModelTC/lightx2v}, 2025.

\bibitem[{Epic Games}(2022)]{ue5}
{Epic Games}.
\newblock Unreal engine 5.
\newblock \url{https://www.unrealengine.com/unreal-engine-5}, 2022.
\newblock Computer software.

\bibitem[Fang et~al.(2026)Fang, He, Tang, Zhang, Li, Liu, Wan, and Gai]{3dmo}
Zhixue Fang, Xu~He, Songlin Tang, Haoxian Zhang, Qingfeng Li, Xiaoqiang Liu,
  Pengfei Wan, and Kun Gai.
\newblock 3d-aware implicit motion control for view-adaptive human video
  generation.
\newblock \emph{arXiv preprint arXiv:2602.03796}, 2026.

\bibitem[Ferguson et~al.(2025)Ferguson, Osman, Bescos, Stoll, Twigg, Lassner,
  Otte, Vignola, Prada, Bogo, Santesteban, Romero, Zarate, Lee, Park, Yang,
  Doublestein, Venkateshan, Kitani, Kavan, Farra, Hu, Cioffi, Fabris, Ranieri,
  Modarres, Kadlecek, Khirodkar, Abdrashitov, Prévost, Rajbhandari, Mallet,
  Pearsall, Kao, Kumar, Parrish, Yu, Saito, Shiratori, Wang, Tung, Xu, Dong,
  Chen, Xu, Ye, and Jiang]{mhr}
Aaron Ferguson, Ahmed A.~A. Osman, Berta Bescos, Carsten Stoll, Chris Twigg,
  Christoph Lassner, David Otte, Eric Vignola, Fabian Prada, Federica Bogo,
  Igor Santesteban, Javier Romero, Jenna Zarate, Jeongseok Lee, Jinhyung Park,
  Jinlong Yang, John Doublestein, Kishore Venkateshan, Kris Kitani, Ladislav
  Kavan, Marco~Dal Farra, Matthew Hu, Matthew Cioffi, Michael Fabris, Michael
  Ranieri, Mohammad Modarres, Petr Kadlecek, Rawal Khirodkar, Rinat
  Abdrashitov, Romain Prévost, Roman Rajbhandari, Ronald Mallet, Russell
  Pearsall, Sandy Kao, Sanjeev Kumar, Scott Parrish, Shoou-I Yu, Shunsuke
  Saito, Takaaki Shiratori, Te-Li Wang, Tony Tung, Yichen Xu, Yuan Dong, Yuhua
  Chen, Yuanlu Xu, Yuting Ye, and Zhongshi Jiang.
\newblock Mhr: Momentum human rig, 2025.
\newblock URL \url{https://arxiv.org/abs/2511.15586}.

\bibitem[{Gemini Team, Google}(2023)]{gemini}
{Gemini Team, Google}.
\newblock Gemini: A family of highly capable multimodal models, 2023.
\newblock URL \url{https://arxiv.org/abs/2312.11805}.

\bibitem[{Google DeepMind}(2025)]{nanobanana}
{Google DeepMind}.
\newblock Nano banana image generation via gemini api.
\newblock \url{https://ai.google.dev/gemini-api/docs/image-generation}, 2025.
\newblock Accessed: 2026-05-20.

\bibitem[Han et~al.(2025)Han, Li, Chen, Yuan, Wu, Leong, Du, Fu, Li, Zhang,
  Zhang, jia Li, and Ni]{videobench}
Hui Han, Siyuan Li, Jiaqi Chen, Yiwen Yuan, Yuling Wu, Chak~Tou Leong, Hanwen
  Du, Junchen Fu, Youhua Li, Jie Zhang, Chi Zhang, Li~jia Li, and Yongxin Ni.
\newblock Video-bench: Human-aligned video generation benchmark, 2025.
\newblock URL \url{https://arxiv.org/abs/2504.04907}.

\bibitem[Hore \& Ziou(2010)Hore and Ziou]{psnr}
Alain Hore and Djemel Ziou.
\newblock Image quality metrics: Psnr vs. ssim.
\newblock In \emph{2010 20th international conference on pattern recognition},
  pp.\  2366--2369. IEEE, 2010.

\bibitem[Hu(2024)]{animateanyone}
Li~Hu.
\newblock Animate anyone: Consistent and controllable image-to-video synthesis
  for character animation.
\newblock In \emph{Proceedings of the IEEE/CVF conference on computer vision
  and pattern recognition}, pp.\  8153--8163, 2024.

\bibitem[Hu et~al.(2025)Hu, Wang, Shen, Gao, Meng, Zhuo, Zhang, Zhang, and
  Bo]{animateanyone2}
Li~Hu, Guangyuan Wang, Zhen Shen, Xin Gao, Dechao Meng, Lian Zhuo, Peng Zhang,
  Bang Zhang, and Liefeng Bo.
\newblock Animate anyone 2: High-fidelity character image animation with
  environment affordance.
\newblock In \emph{Proceedings of the IEEE/CVF International Conference on
  Computer Vision}, pp.\  10207--10217, 2025.

\bibitem[Hu et~al.(2026)Hu, Gong, Yang, An, Xu, and Liu]{hu2026multianimate}
Yingcheng Hu, Haowen Gong, Chuanguang Yang, Zhulin An, Yongjun Xu, and Songhua
  Liu.
\newblock Multianimate: Pose-guided image animation made extensible.
\newblock \emph{arXiv preprint arXiv:2602.21581}, 2026.

\bibitem[Jiang et~al.(2025)Jiang, Han, Mao, Zhang, Pan, and Liu]{vace}
Zeyinzi Jiang, Zhen Han, Chaojie Mao, Jingfeng Zhang, Yulin Pan, and Yu~Liu.
\newblock Vace: All-in-one video creation and editing.
\newblock In \emph{Proceedings of the IEEE/CVF International Conference on
  Computer Vision}, pp.\  17191--17202, 2025.

\bibitem[Li et~al.(2026)Li, Gao, Hassan, Feng, Pan, Luan, and
  Alahi]{everanimate}
Wuyang Li, Yang Gao, Mariam Hassan, Lan Feng, Wentao Pan, Po-Chien Luan, and
  Alexandre Alahi.
\newblock Everanimate: Minute-scale human animation via latent flow
  restoration, 2026.
\newblock URL \url{https://arxiv.org/abs/2605.15042}.

\bibitem[Liang et~al.(2026)Liang, He, Wang, Liao, Zhang, Chen, and
  Yuan]{sdpose}
Shuang Liang, Jing He, Chuanmeizhi Wang, Lejun Liao, Guo Zhang, Yingcong Chen,
  and Yuan Yuan.
\newblock Sdpose: Exploiting diffusion priors for out-of-domain and robust pose
  estimation, 2026.
\newblock URL \url{https://arxiv.org/abs/2509.24980}.

\bibitem[Luo et~al.(2026)Luo, Liang, Rong, Luo, Hu, Hou, Chang, Li, Zhang, and
  Gao]{dreamactor-m2}
Mingshuang Luo, Shuang Liang, Zhengkun Rong, Yuxuan Luo, Tianshu Hu, Ruibing
  Hou, Hong Chang, Yong Li, Yuan Zhang, and Mingyuan Gao.
\newblock Dreamactor-m2: Universal character image animation via spatiotemporal
  in-context learning.
\newblock \emph{arXiv preprint arXiv:2601.21716}, 2026.

\bibitem[Ma et~al.(2026{\natexlab{a}})Ma, Liu, Zhu, Yang, Feng, Zhang, Yan, Li,
  Han, Qi, and Chen]{followyourmotion}
Yue Ma, Yulong Liu, Qiyuan Zhu, Ayden Yang, Kunyu Feng, Xinhua Zhang, Zexuan
  Yan, Zhifeng Li, Sirui Han, Chenyang Qi, and Qifeng Chen.
\newblock Follow-your-motion: Video motion transfer via efficient
  spatial-temporal decoupled finetuning, 2026{\natexlab{a}}.
\newblock URL \url{https://arxiv.org/abs/2506.05207}.

\bibitem[Ma et~al.(2026{\natexlab{b}})Ma, Wang, Ren, Zheng, Liu, Guo, Feng,
  Xue, Zhao, Schindler, Chen, and Zhang]{fastvmt}
Yue Ma, Zhikai Wang, Tianhao Ren, Mingzhe Zheng, Hongyu Liu, Jiayi Guo, Kunyu
  Feng, Yuxuan Xue, Zixiang Zhao, Konrad Schindler, Qifeng Chen, and Linfeng
  Zhang.
\newblock Fastvmt: Eliminating redundancy in video motion transfer,
  2026{\natexlab{b}}.
\newblock URL \url{https://arxiv.org/abs/2602.05551}.

\bibitem[S{\'a}r{\'a}ndi \& Pons-Moll(2024)S{\'a}r{\'a}ndi and Pons-Moll]{nlf}
Istv{\'a}n S{\'a}r{\'a}ndi and Gerard Pons-Moll.
\newblock Neural localizer fields for continuous 3d human pose and shape
  estimation.
\newblock \emph{Advances in Neural Information Processing Systems},
  37:\penalty0 140032--140065, 2024.

\bibitem[Shi et~al.(2025)Shi, Xu, Li, Peng, Yang, Lu, Hu, and Zhang]{onetoall}
Shijun Shi, Jing Xu, Zhihang Li, Chunli Peng, Xiaoda Yang, Lijing Lu, Kai Hu,
  and Jiangning Zhang.
\newblock One-to-all animation: Alignment-free character animation and image
  pose transfer.
\newblock \emph{arXiv preprint arXiv:2511.22940}, 2025.

\bibitem[Song et~al.(2025)Song, Xu, Zhao, Xie, Gu, Li, Zhang, and
  Luo]{xunimotion}
Guoxian Song, Hongyi Xu, Xiaochen Zhao, You Xie, Tianpei Gu, Zenan Li, Chenxu
  Zhang, and Linjie Luo.
\newblock X-unimotion: Animating human images with expressive, unified and
  identity-agnostic motion latents.
\newblock In \emph{Proceedings of the SIGGRAPH Asia 2025 Conference Papers},
  pp.\  1--11, 2025.

\bibitem[Tan et~al.(2025)Tan, Gong, Wang, Zhang, Zheng, Zheng, Zheng, Chen, and
  Yang]{animatex}
Shuai Tan, Biao Gong, Xiang Wang, Shiwei Zhang, DanDan Zheng, Ruobing Zheng,
  Kecheng Zheng, Jingdong Chen, and Ming Yang.
\newblock Animate-x: Universal character image animation with enhanced motion
  representation.
\newblock In \emph{The Thirteenth International Conference on Learning
  Representations}, 2025.
\newblock URL \url{https://openreview.net/forum?id=1IuwdOI4Zb}.

\bibitem[Team et~al.(2026)Team, Chen, Ding, Fang, Gai, He, He, Hua, Lao, Li,
  Liu, Liu, Liu, Shi, Shi, Sun, Tang, Wan, Wen, Wu, Zhang, Zhao, Zhang, and
  Zhou]{klingmotion}
Kling Team, Jialu Chen, Yikang Ding, Zhixue Fang, Kun Gai, Kang He, Xu~He,
  Jingyun Hua, Mingming Lao, Xiaohan Li, Hui Liu, Jiwen Liu, Xiaoqiang Liu, Fan
  Shi, Xiaoyu Shi, Peiqin Sun, Songlin Tang, Pengfei Wan, Tiancheng Wen,
  Zhiyong Wu, Haoxian Zhang, Runze Zhao, Yuanxing Zhang, and Yan Zhou.
\newblock Kling-motioncontrol technical report, 2026.
\newblock URL \url{https://arxiv.org/abs/2603.03160}.

\bibitem[Unterthiner et~al.(2018)Unterthiner, Van~Steenkiste, Kurach, Marinier,
  Michalski, and Gelly]{fvd}
Thomas Unterthiner, Sjoerd Van~Steenkiste, Karol Kurach, Raphael Marinier,
  Marcin Michalski, and Sylvain Gelly.
\newblock Towards accurate generative models of video: A new metric \&
  challenges.
\newblock \emph{arXiv preprint arXiv:1812.01717}, 2018.

\bibitem[Wallace et~al.(2024)Wallace, Dang, Rafailov, Zhou, Lou, Purushwalkam,
  Ermon, Xiong, Joty, and Naik]{Wallace_2024_diffusion_dpo}
Bram Wallace, Meihua Dang, Rafael Rafailov, Linqi Zhou, Aaron Lou, Senthil
  Purushwalkam, Stefano Ermon, Caiming Xiong, Shafiq Joty, and Nikhil Naik.
\newblock Diffusion model alignment using direct preference optimization.
\newblock In \emph{Proceedings of the IEEE/CVF Conference on Computer Vision
  and Pattern Recognition (CVPR)}, pp.\  8228--8238, June 2024.

\bibitem[Wan et~al.(2025)Wan, Wang, Ai, Wen, Mao, Xie, Chen, Yu, Zhao, Yang,
  et~al.]{wan}
Team Wan, Ang Wang, Baole Ai, Bin Wen, Chaojie Mao, Chen-Wei Xie, Di~Chen,
  Feiwu Yu, Haiming Zhao, Jianxiao Yang, et~al.
\newblock Wan: Open and advanced large-scale video generative models.
\newblock \emph{arXiv preprint arXiv:2503.20314}, 2025.

\bibitem[Wang et~al.(2025)Wang, Zhang, Tang, Zhang, Gao, Wang, and
  Sang]{unianimate-dit}
Xiang Wang, Shiwei Zhang, Longxiang Tang, Yingya Zhang, Changxin Gao, Yuehuan
  Wang, and Nong Sang.
\newblock Unianimate-dit: Human image animation with large-scale video
  diffusion transformer, 2025.
\newblock URL \url{https://arxiv.org/abs/2504.11289}.

\bibitem[Wang et~al.(2004)Wang, Bovik, Sheikh, and Simoncelli]{ssim}
Zhou Wang, Alan~C Bovik, Hamid~R Sheikh, and Eero~P Simoncelli.
\newblock Image quality assessment: from error visibility to structural
  similarity.
\newblock \emph{IEEE transactions on image processing}, 13\penalty0
  (4):\penalty0 600--612, 2004.

\bibitem[Xu et~al.(2026{\natexlab{a}})Xu, Zheng, Wang, Yu, Chen, Zhang, Zhou,
  Lee, Li, and Jiang]{hypermotion}
Shuolin Xu, Siming Zheng, Ziyi Wang, HC~Yu, Jinwei Chen, Huaqi Zhang, Daquan
  Zhou, Tong-Yee Lee, Bo~Li, and Peng-Tao Jiang.
\newblock Hypermotionx: The dataset and benchmark with dit-based pose-guided
  human image animation of complex motions, 2026{\natexlab{a}}.
\newblock URL \url{https://arxiv.org/abs/2505.22977}.

\bibitem[Xu et~al.(2022)Xu, Zhang, Zhang, and Tao]{xu2022vitpose}
Yufei Xu, Jing Zhang, Qiming Zhang, and Dacheng Tao.
\newblock Vi{TP}ose: Simple vision transformer baselines for human pose
  estimation.
\newblock In \emph{Advances in Neural Information Processing Systems}, 2022.

\bibitem[Xu et~al.(2026{\natexlab{b}})Xu, Ma, Wang, Peng, Liang, and Li]{mocha}
Zhengbo Xu, Jie Ma, Ziheng Wang, Zhan Peng, Jun Liang, and Jing Li.
\newblock End-to-end video character replacement without structural guidance.
\newblock \emph{arXiv preprint arXiv:2601.08587}, 2026{\natexlab{b}}.

\bibitem[Yan et~al.(2025)Yan, Ye, Yang, Teng, Dong, Wen, Gu, Liu, and
  Tang]{scail}
Wenhao Yan, Sheng Ye, Zhuoyi Yang, Jiayan Teng, ZhenHui Dong, Kairui Wen,
  Xiaotao Gu, Yong-Jin Liu, and Jie Tang.
\newblock Scail: Towards studio-grade character animation via in-context
  learning of 3d-consistent pose representations.
\newblock \emph{arXiv preprint arXiv:2512.05905}, 2025.

\bibitem[Yang et~al.(2026)Yang, Kukreja, Pinkus, Sagar, Fan, Park, Shin, Cao,
  Liu, Ugrinovic, Feiszli, Malik, Dollar, and Kitani]{sam3db}
Xitong Yang, Devansh Kukreja, Don Pinkus, Anushka Sagar, Taosha Fan, Jinhyung
  Park, Soyong Shin, Jinkun Cao, Jiawei Liu, Nicolas Ugrinovic, Matt Feiszli,
  Jitendra Malik, Piotr Dollar, and Kris Kitani.
\newblock Sam 3d body: Robust full-body human mesh recovery, 2026.
\newblock URL \url{https://arxiv.org/abs/2602.15989}.

\bibitem[Yang et~al.(2023)Yang, Zeng, Yuan, and Li]{yang2023effectivedwpose}
Zhendong Yang, Ailing Zeng, Chun Yuan, and Yu~Li.
\newblock Effective whole-body pose estimation with two-stages distillation.
\newblock In \emph{Proceedings of the IEEE/CVF International Conference on
  Computer Vision}, pp.\  4210--4220, 2023.

\bibitem[Yang et~al.(2025)Yang, Teng, Zheng, Ding, Huang, Xu, Yang, Hong,
  Zhang, Feng, et~al.]{cogvideox}
Zhuoyi Yang, Jiayan Teng, Wendi Zheng, Ming Ding, Shiyu Huang, Jiazheng Xu,
  Yuanming Yang, Wenyi Hong, Xiaohan Zhang, Guanyu Feng, et~al.
\newblock Cogvideox: Text-to-video diffusion models with an expert transformer.
\newblock In \emph{International Conference on Learning Representations},
  volume 2025, pp.\  83048--83077, 2025.

\bibitem[Zhang et~al.(2026)Zhang, Cao, Li, Zhao, Cui, Hou, Wu, Chen, Xu, Wang,
  and Ma]{steadydancer}
Jiaming Zhang, Shengming Cao, Rui Li, Xiaotong Zhao, Yutao Cui, Xinglin Hou,
  Gangshan Wu, Haolan Chen, Yu~Xu, Limin Wang, and Kai Ma.
\newblock Steadydancer: Harmonized and coherent human image animation with
  first-frame preservation, 2026.
\newblock URL \url{https://arxiv.org/abs/2511.19320}.

\bibitem[Zhang et~al.(2018)Zhang, Isola, Efros, Shechtman, and Wang]{lpips}
Richard Zhang, Phillip Isola, Alexei~A Efros, Eli Shechtman, and Oliver Wang.
\newblock The unreasonable effectiveness of deep features as a perceptual
  metric.
\newblock In \emph{Proceedings of the IEEE conference on computer vision and
  pattern recognition}, pp.\  586--595, 2018.

\bibitem[Zhao et~al.(2023)Zhao, Gu, Varma, Luo, Huang, Xu, Wright, Shojanazeri,
  Puglia, Chen, et~al.]{zhao2023pytorch}
Yanli Zhao, Andrew Gu, Rohan Varma, Liang Luo, Chien-Chin Huang, Min Xu, Tengyu
  Wright, Hamid Shojanazeri, Myle Puglia, Soumith Chen, et~al.
\newblock Pytorch fsdp: experiences on scaling fully sharded data parallel.
\newblock \emph{Proceedings of the VLDB Endowment}, 16\penalty0 (12):\penalty0
  3848--3860, 2023.

\bibitem[Zhu et~al.(2024)Zhu, Chen, Dai, Dong, Xu, Cao, Yao, Zhu, and
  Zhu]{champ}
Shenhao Zhu, Junming~Leo Chen, Zuozhuo Dai, Zilong Dong, Yinghui Xu, Xun Cao,
  Yao Yao, Hao Zhu, and Siyu Zhu.
\newblock Champ: Controllable and consistent human image animation with 3d
  parametric guidance.
\newblock In \emph{European Conference on Computer Vision}, pp.\  145--162.
  Springer, 2024.

\end{thebibliography}
  \bibliographystyle{iclr2026_conference}
\fi

\clearpage
\appendix
\begin{center}
    \LARGE
    \textbf{Supplementary Material}\\
    \vspace{0.5em}
    \Large
    \textbf{Appendix}\\
    \vspace{1.0em}
\end{center}

\section{Details on the MotionPair-60K Dataset}\label{app:data-details}

\noindent\textbf{Data Composition.}\enspace We collect source videos from the same source datasets as SCAIL~\citep{scail}. LightX2V~\citep{lightx2v} is adopted for the generators to accelerate synthetic-data generation. The full pipeline yields \textbf{MotionPair-60K}, comprising $59{,}376$ end-to-end motion-transfer pairs with animation and replacement data in a ratio of approximately 3:1. The final data composition and sampling ratios are shown in Tab.~\ref{tab:data_composition}.

\noindent\textbf{Augmentation and Sampling Details.}\enspace We apply random augmentations to the driving video in animation mode. In training, we treat pose-driven animation as a special case of end-to-end driving and randomly sample pose-driven pairs to improve data diversity. We adopt random cropping and stretching for end-to-end animation driving sequences, and random skeleton scaling for pose conditions.

\begin{figure}[h]
    \centering
    \includegraphics[width=0.65\textwidth]{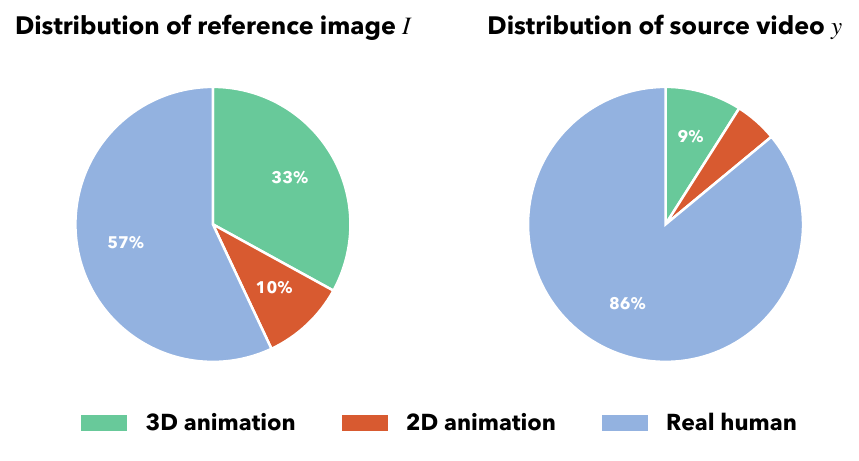}
    \caption{Distribution of data source.}
    \label{fig:data_composition}
\end{figure}

To further improve data diversity, we sample approximately 5\% of the motion pairs without reversing the driving direction and directly use the synthetic video $\tilde{\boldsymbol{y}}$ as the denoising target.

\begin{table}[h]
\centering
\small
\caption{Composition of \textbf{MotionPair-60K}, along with the additional pose-driven dataset, and their corresponding sampling ratios used during training.}
\label{tab:data_composition}
\begin{tabular}{@{}l l r c@{}}
\toprule
Source & Construction & \#Pairs & Sampling Ratio \\
\midrule
SCAIL        & Single animation & $31{,}895$ & \multirow{2}{*}{$60\%$} \\
Wan-Animate  & Single animation & $13{,}847$ & \\
\midrule
\multirow{2}{*}{MoCha} & Single replacement & $9{,}249$ & \multirow{2}{*}{$20\%$} \\
                       & Multi replacement  & $4{,}385$ & \\
\midrule
- & Single/Multi Pose Extraction & $\sim 100{,}000$ & $20\%$ \\
\bottomrule
\end{tabular}
\end{table}

\section{End-to-End Training Details}\label{app:implementation-details}
\noindent\textbf{Long Video Generation.}\enspace Following Wan-Animate, during training, we randomly replace the first two latent frames with conditional history latents, enabling the model to autoregressively extend videos beyond the training clip length.

\noindent\textbf{Implementation Details.}\enspace During end-to-end training, we fully fine-tune the backbone for 3{,}500 steps with a batch size of 128 and a learning rate of $10^{-5}$. We train with spatial resolutions ranging from $512{\times}896$ to $704{\times}1280$ and clip lengths of either 65 or 81 frames. Training is conducted on 64 NVIDIA H100 GPUs for approximately one week using FSDP-2~\citep{zhao2023pytorch}.

\section{Post-Training Details}\label{app:post-training-details}


\begin{figure*}[htbp]
    \centering
    \includegraphics[width=0.935\textwidth]{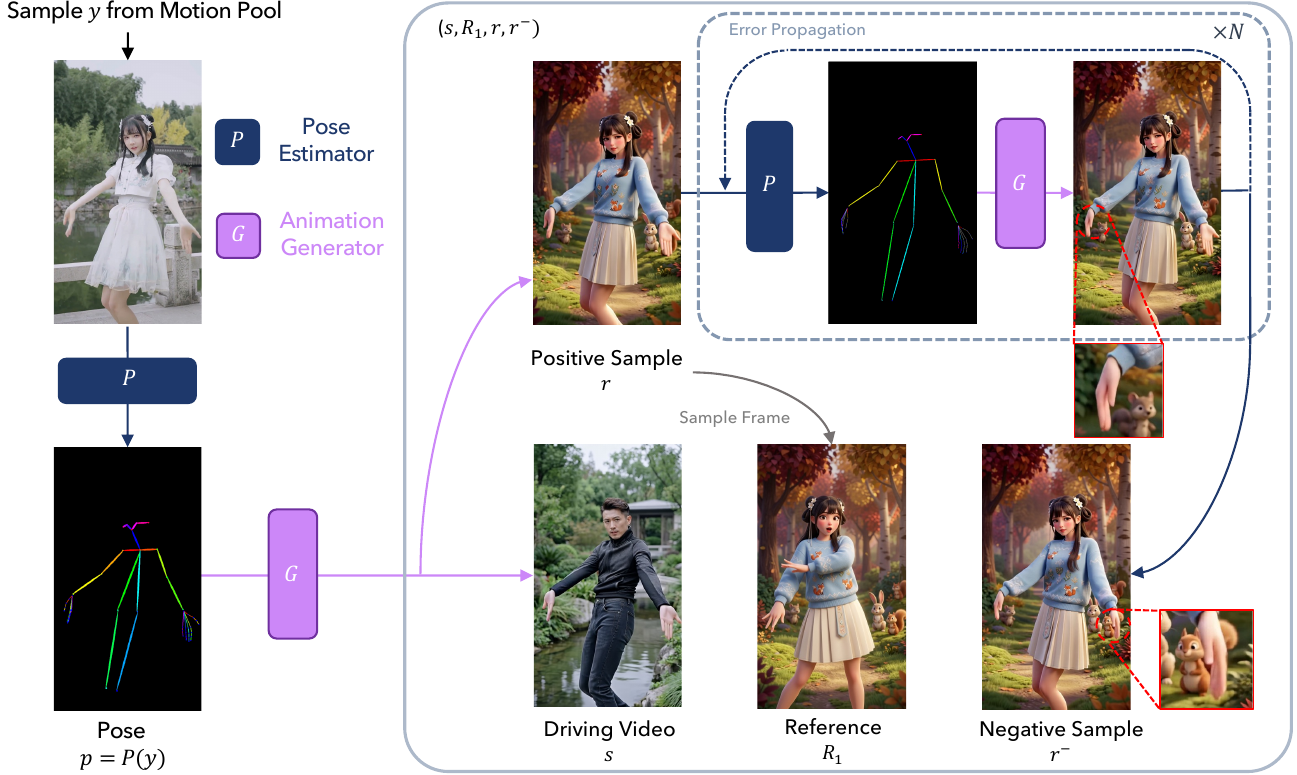}
    \caption{The overview of the pipeline for constructing preference data in Bias-Aware DPO training.}
    \label{fig:dpo_data_pipeline}
\end{figure*}

\subsection{Details of Preference Dataset}

We visualize the construction process in Fig.~\ref{fig:dpo_data_pipeline}. Unlike the end-to-end training stage, where ground-truth videos are used in a reverse-driving manner, preference optimization directly treats the synthesized video $r$ as the preferred target, so its fidelity directly affects the optimization signal. We therefore use the accurate SDPose estimator~\citep{sdpose} to generate clean positive samples and apply strict filtering. To amplify error propagation in negative samples, we use the less accurate ViTPose~\citep{xu2022vitpose} for the additional extraction passes $P'$ or $P''$ in Eq.~\eqref{eq:negative-propagation}. Since both branches share the same reference image $R$ and general
global motion, the resulting gap between $r$ and $r^{-}$ is concentrated in fine-grained articulation rather than in identity or global motion. For the animation generator $\mathcal{G}$ used in post-training we adopt Wan-Animate, which is also used to synthesize close-up shots as end-to-end training pairs. The final pipeline yields around 1K pairs of preference data.

\subsection{Bias-Aware DPO Implementation}

\noindent\textbf{General Formulation.}\enspace Following Diffusion-DPO~\citep{Wallace_2024_diffusion_dpo}, we optimize a
trainable flow matching model $v_\theta$ against a frozen reference model
$v_{\mathrm{ref}}$. Given a preference dataset
$\mathcal D=\{(x_0^+,x_0^-)\}$ consisting of preferred and dispreferred samples, the DPO objective is
\begin{align}\label{eq:dpo}
    \mathcal{L}_{\text{DPO}} = & - \mathbb{E}_{\left(x_0^+, x_0^-\right) \sim \mathcal{D}, \, x_1 \sim \mathcal{N}(0,\mathbf{I}), \, t } \notag \\ &
    \left[ \log \sigma\! \left( - \frac{\beta}{2} \left( \Delta(x_0^+, x_1, t) - \Delta(x_0^-, x_1, t) \right) \right) \right],
\end{align}
where $x_0^+$ and $x_0^-$ denote the preferred and dispreferred samples,
respectively.
The term $\Delta(\cdot, \cdot, \cdot)$ measures the relative flow-matching error between
the trainable model and the frozen reference model.
\begin{align}
    \Delta(x_0, x_1, t)  & =  \| v_{\theta}(x_t, t) - v \|_2^2 -
    \| v_{\text{ref}}(x_t, t) - v \|_2^2 \\
    x_t & = tx_1 + (1-t)x_0 \\
    v & = x_1 - x_0
\end{align}

\noindent\textbf{Regional DPO.}\enspace As described in the main paper, the difference of the preference pairs comes from the estimation error of pose estimators, which is most significant in hand movements. To highlight such difference and avoid distraction brought by other factors, we apply the DPO objective only within hand regions.

Specifically, given a hand mask $M$, the per-sample DPO score is computed using masked velocity prediction errors:
\begin{align}
    \Delta_M(x_0,x_1,t)
    =
    \| M \odot (v_{\theta}(x_t,t)-v)\|_2^2
    - \notag \\
    \| M \odot (v_{\mathrm{ref}}(x_t,t)-v)\|_2^2,
\end{align}
where $\odot$ denotes element-wise multiplication. We design $M$ as the union of hand bounding boxes of positive and negative samples at each frame, and directly downsample it into a mask in latent space.

We replace $\Delta$ in Eq.~(\ref{eq:dpo}) with $\Delta_M$ when computing the DPO loss, and the DPO term over a preference pair with regional mask becomes
\begin{align}
    & \mathcal{L}_{\mathrm{DPO}}^{M}(x_0^+, x_0^-, x_1, t) \notag \\
    = \, &
    - \log \sigma \! \left( - \frac{\beta}{2} \left( \Delta_M(x_0^+, x_1, t) - \Delta_M(x_0^-, x_1, t) \right) \right).
\end{align}

\noindent\textbf{SFT Anchor.}\enspace Optimizing the DPO objective alone leads to unstable training. We adopt a common approach, adding a supervised fine-tuning (SFT) item over positive samples to mitigate the problem:

\begin{align}
    \mathcal{L}_{\text{SFT}}(x_0^+, x_1, t) & =
    \| v_{\theta}(x_t^{+}, t) - v^+ \|_2^2 \\
    x_t^{+} & = tx_1 + (1-t)x_0^+ \\
    v^+ & = x_1 - x_0^+
\end{align}

To stabilize training, we jointly optimize the SFT objective with the DPO objective:
\begin{align}
    \mathcal{L}
    = &\, \mathbb{E}_{\left(x_0^+, x_0^-, M\right) \sim \mathcal{D}, \, x_1 \sim \mathcal{N}(0,\mathbf{I}), \, t} \notag \\
    & \left[
    \mathcal{L}_{\mathrm{SFT}}(x_0^+, x_1, t)
    +
    \lambda
    \mathcal{L}_{\mathrm{DPO}}^{M}(x_0^+, x_0^-, x_1, t) \right],
\end{align}
where $\lambda = 0.01$ is used to balance the scales of the two objectives. The SFT term serves as an anchor that  prevents excessive divergence during preference optimization.

\noindent\textbf{Implementation Details.}\enspace During post-training, we freeze the backbone parameters and optimize only LoRA adapters inserted into the transformer layers, with rank $128$. The DPO temperature parameter is set to $\beta= 5000$. The LoRA adapters are tuned for 400 steps, using a learning rate of $1\times10^{-4}$ and a batch size of $24$.

\section{Evaluation Details}
\noindent\textbf{Pose-Driven Metrics.}\enspace For the \textit{self-driven} split, we employed several widely-used quantitative metrics, including PSNR~\citep{psnr}, SSIM~\citep{ssim}, LPIPS~\citep{lpips}, and FVD~\citep{fvd}.

\noindent\textbf{Human Evaluation Metrics.}\enspace To evaluate the generated results in cross-identity settings, human evaluation is still the most convincing method. To conduct reasonable subjective studies, we design several metrics:

(1) \textit{Motion Accuracy}, which measures how faithfully the generated motion follows the driving signal in a frame-by-frame manner.

(2) \textit{Identity Consistency}, measuring the consistency of the subject’s appearance with the reference image.

For single character image animation, we measure:

(3) \textit{Physical Plausibility}, assessing whether the generated motions comply with basic physical constraints such as gravity, support, and momentum conservation, especially when object interactions are involved. This metric penalizes unrealistic behaviors like hovering in midair, objects morphing, or objects penetrating into human body.

For multi character image animation, we measure:

(4) \textit{Identity Isolation}, making sure that one character's limbs do not unnaturally merge with another character's body, and their clothing remains strictly separated.

For replacement scenarios, we measure:

(5) \textit{Environment Integration}. This metric evaluates whether the newly replaced characters fit naturally into the original scene and how well the reference video's environment is maintained in the generated output, including the consistency of the background and the preservation of character-object interactions.

\noindent\textbf{Automatic Metrics.}\enspace For cases where the original quality of the video can work as a strong indicator of the performance, we adopt VideoBench’s human-aligned automatic protocol following DreamActor-M2~\citep{dreamactor-m2}, focusing on three key perceptual dimensions: Imaging Quality, Temporal Consistency, and Appearance Consistency. The protocol evaluates all three dimensions on a 1–5 scale (1=very poor, 2=poor, 3=moderate, 4=good, 5=excellent).

\noindent\textbf{Studio-Bench Details.}\enspace The self-driven single-character split of Studio-Bench contains 80 cases while the cross-identity split contains 60/30/28 cases for single-character animation, multi-character animation and single-character replacement, respectively. All evaluation cases are excluded from the training data. For each cross-identity case, multiple independent raters select one of three outcomes:
Good (our method is better), Same (the two methods are
comparable), or Bad (the competing method is better).
Let $n_G$ and $n_B$ denote the numbers of Good and Bad
votes, respectively. We assign the case-level outcome as
Good if $n_G > n_B$, Bad if $n_B > n_G$, and Same if
$n_G = n_B$. Each case is evaluated by at least five independent raters under a blind evaluation protocol.

We provide visualizations of the representations used by both our method and the baselines under the evaluation of the self-driven split of Studio-Bench. For the SAM3D-Body Mesh~\citep{sam3db} adopted in Tab.~\ref{tab:more-baselines}, we use the standard gray mesh in the MHR format~\citep{mhr}, which provides visually richer information than skeletal representations by capturing detailed spatial structure of limbs. For each pose-based method, we adopt its default pose estimator~\citep{nlf, yang2023effectivedwpose, xu2022vitpose} and the corresponding skeleton representation.

\begin{figure}[h]
    \centering
    \includegraphics[width=0.70\textwidth]{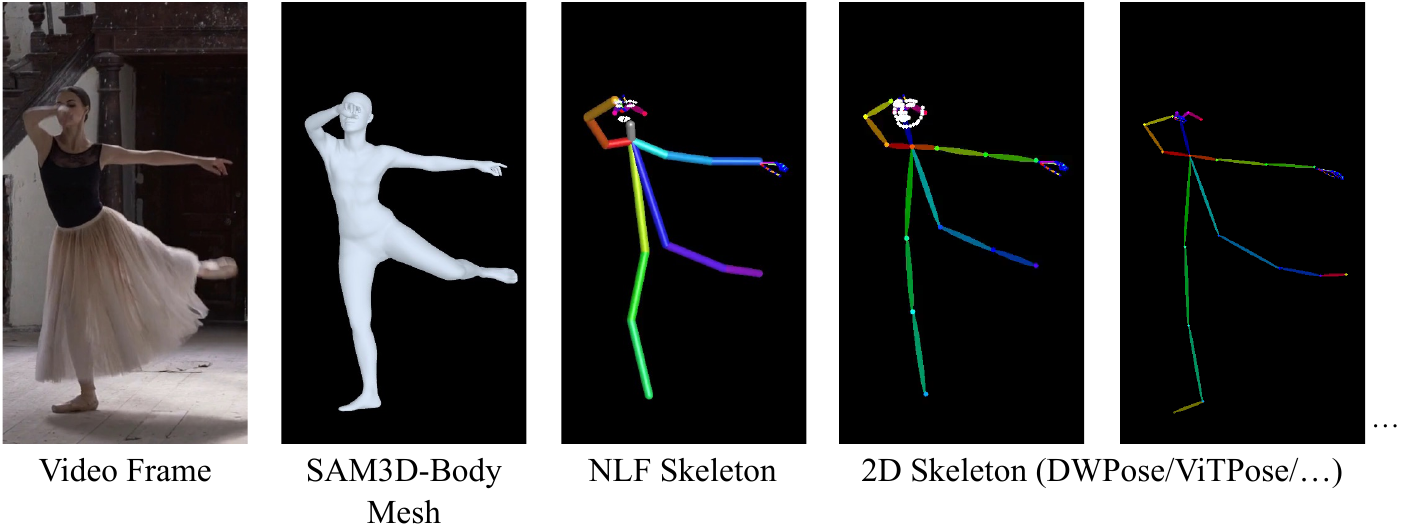}
    \caption{Visualizations of representations in evaluation.}
    \label{fig:pose_comparision}
\end{figure}

\section{More Examples}
\begin{figure}[H]
    \centering
    \includegraphics[width=\textwidth]{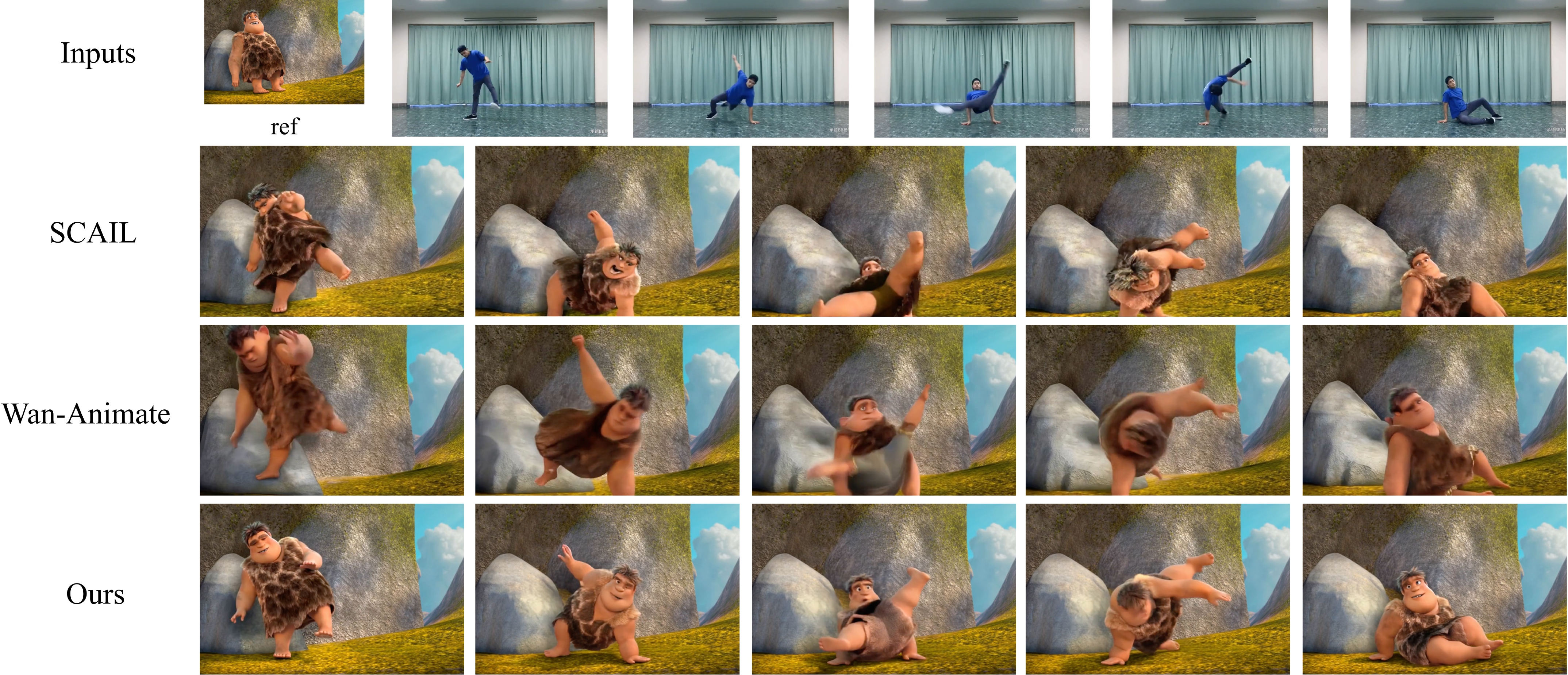}
    \caption{Examples of complex cross-body-shape character image animation. Our method maintains decent character consistency under complex motions.}
    \label{fig:finalcomp1}
\end{figure}

\begin{figure}[H]
    \centering
    \includegraphics[width=\textwidth]{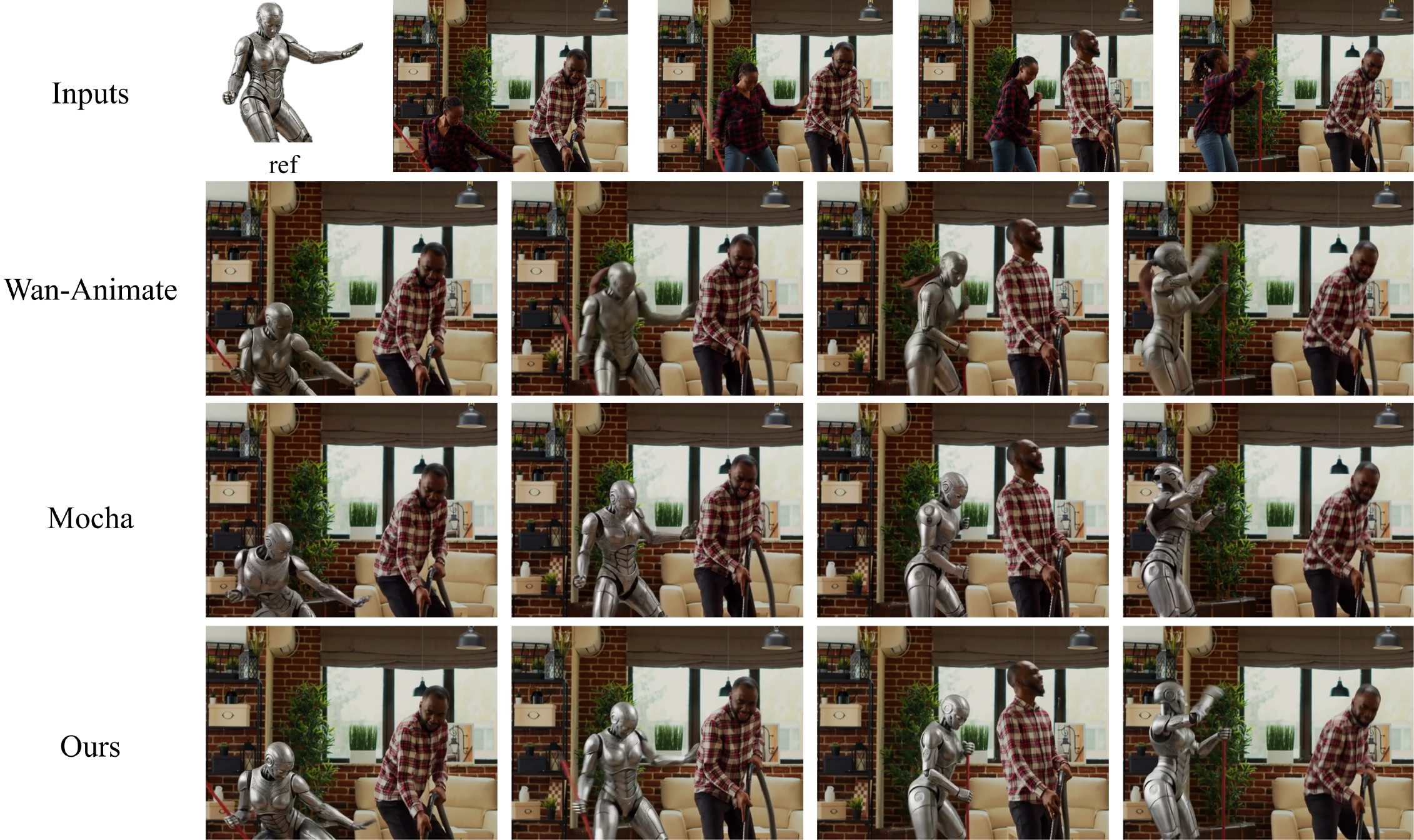}
    \caption{Examples requiring fine-grained HOI. Our method simultaneously preserves correct character identity and fine-grained objects (e.g., thin sticks) during interaction. Zoom-in for better details.}
    \label{fig:finalcomp0}
\end{figure}

\begin{figure}[H]
    \centering
    \includegraphics[width=\textwidth]{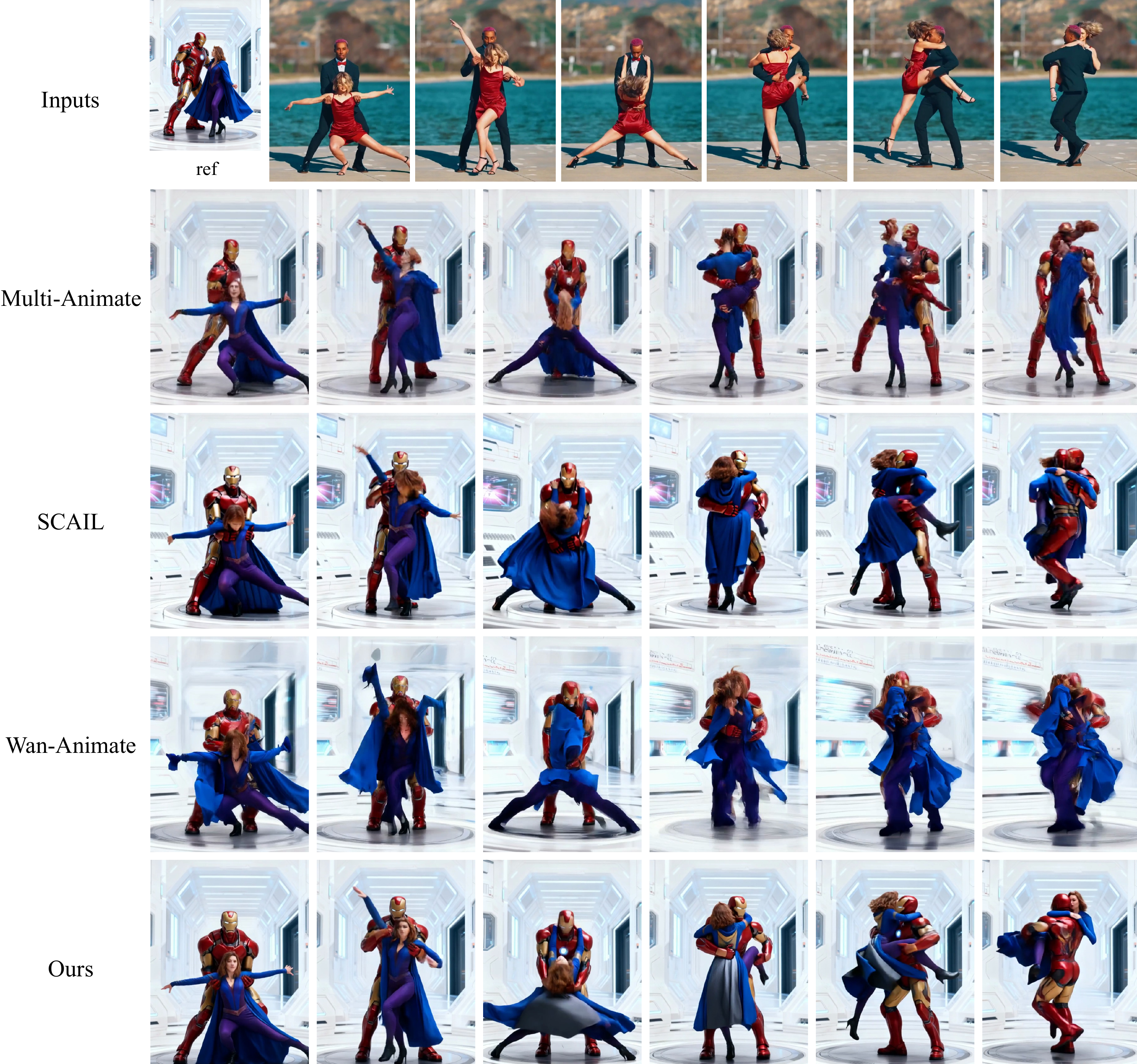}
    \caption{Examples of complex multi-character interactions. Our method accurately captures the interaction relationships among multiple characters with proper identity isolation.}
    \label{fig:finalcomp2}
\end{figure}

We provide additional visualization results to further demonstrate the generalization of SCAIL-2. Our method remains robust under challenging scenarios
in Fig.~\ref{fig:finalcomp1}, Fig.~\ref{fig:finalcomp0} and Fig.~\ref{fig:finalcomp2}. Detailed video demonstrations are provided on our project page.

\section{Release and Licensing}
To support reproducibility, we will open-source the inference code and model weights and release a large subset of MotionPair-60K. The code will be released under the Apache License 2.0, while the released dataset subset will be distributed under terms consistent with the licenses, privacy requirements, and redistribution restrictions of its source data.

\section{Ethical Considerations}
Character animation and replacement can support creative production, but may also enable deceptive or non-consensual impersonation. SCAIL-2 is intended for research and authorized creative use; users should obtain the necessary rights and consent for input identities and media and disclose generated or edited content when appropriate. Future releases will consider safeguards such as provenance metadata, watermarking, and misuse detection.

\end{document}